\documentclass{article}

\usepackage{iclr2026_conference,times}

\usepackage{amsmath,amsfonts,bm}

\def\eqref#1{equation~\ref{#1}}

\def\1{\bm{1}}

\DeclareMathAlphabet{\mathsfit}{\encodingdefault}{\sfdefault}{m}{sl}
\SetMathAlphabet{\mathsfit}{bold}{\encodingdefault}{\sfdefault}{bx}{n}

\usepackage{float}
\usepackage{url}
\usepackage{booktabs}
\usepackage{graphicx}
\usepackage{amsmath,amssymb}
\usepackage[utf8]{inputenc}
\usepackage{xcolor}
\usepackage[colorinlistoftodos,textsize=small]{todonotes}

\newcommand{\unispaceheader}{%
  \raisebox{-0.16\height}{%
    \includegraphics[height=0.21in]{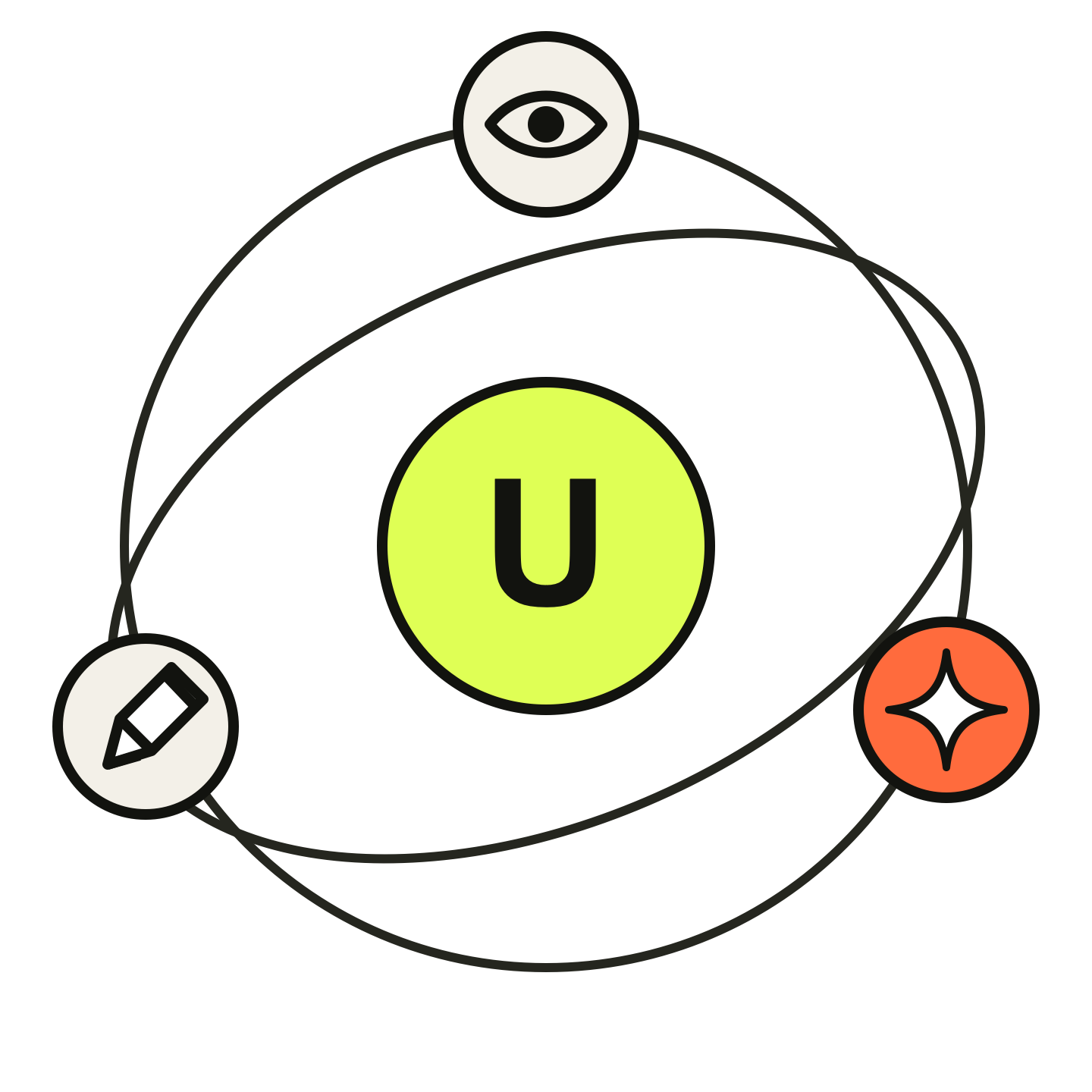}%
  }%
  \hspace{0.45em}%
  {\sffamily\bfseries\small UniSpace}%
}
\usepackage{hyperref}
\hypersetup{
  colorlinks=true,
  linkcolor=blue,
  urlcolor=blue,
  citecolor=blue
}

\iclrfinalcopy

\title{UniSpace: Unified Visual Representation and\\
Scalable Multimodal Modeling}

\author{%
  Jinbo Yan
  \qquad
  Limeng Qiao
  \qquad
  Jie Qin
  \qquad
  Junyan He
  \\[0.4em]
  Feize Wu
  \qquad
  Guanglu Wan
  \\[0.8em]
  {\large\color{black}Meituan}
}

\makeatletter
\def\@maketitle{%
  \begin{center}

    {\huge\bfseries\sffamily
      \linespread{1.08}\selectfont
      \@title\par
    }

    \vskip 0.38in

    {\Large
      \linespread{1.25}\selectfont
      \@author\par
    }

    \vskip 0.28in

  \end{center}
}
\makeatother

\begin{document}

\maketitle
\fancyhead{}
\fancyhead[C]{\unispaceheader}

\begin{abstract}
Semantic vision encoders have become a central visual interface for multimodal
understanding and semantic conditioning in image generation. However, their
final tokens discard fine-grained visual details, leading to poor pixel
reconstruction and limiting their use in reconstruction-sensitive tasks such as
image generation and editing. 
In this work, we ask whether understanding, generation, and editing can be
modeled in a single visual representation space built from a pretrained
semantic ViT. We show that the frozen Transformer blocks of a semantic ViT are
not intrinsically unable to preserve visual details. Instead, the original patch
parameterization drives the representation toward semantic abstraction, making
fine-grained information difficult to recover from the final tokens. Based on
this observation, we introduce \emph{Patch Reparameterization}, which preserves
the original semantic pathway while adding a reconstruction-aware patch embedding
that provides fine-grained visual information to the same frozen ViT blocks.
The resulting unified representation preserves multimodal understanding while
enabling high-fidelity image reconstruction and a favorable
reconstruction--generation trade-off.
We further scale this representation into \emph{UniSpace}, an 8B
Mixture-of-Transformer-Experts model that performs understanding, generation,
and editing in the same visual space without a separate VAE pathway.
System-level evaluations demonstrate practical text-to-image generation and
instruction-based image editing, showing that a reparameterized pretrained ViT
can serve as a unified visual interface for scalable multimodal modeling.
\vspace{10pt}

\noindent\textbf{Project}: \href{https://yjb6.github.io/UniSpace}{https://yjb6.github.io/UniSpace/} \\
\textbf{Hugging Face}: \href{https://huggingface.co/yjb6/UniSpace}{https://huggingface.co/yjb6/UniSpace} \\
\textbf{GitHub}: \href{https://github.com/yjb6/UniSpace}{https://github.com/yjb6/UniSpace}

\vspace{20pt}
\begin{figure}[h]
\centering

\includegraphics[width=\textwidth]{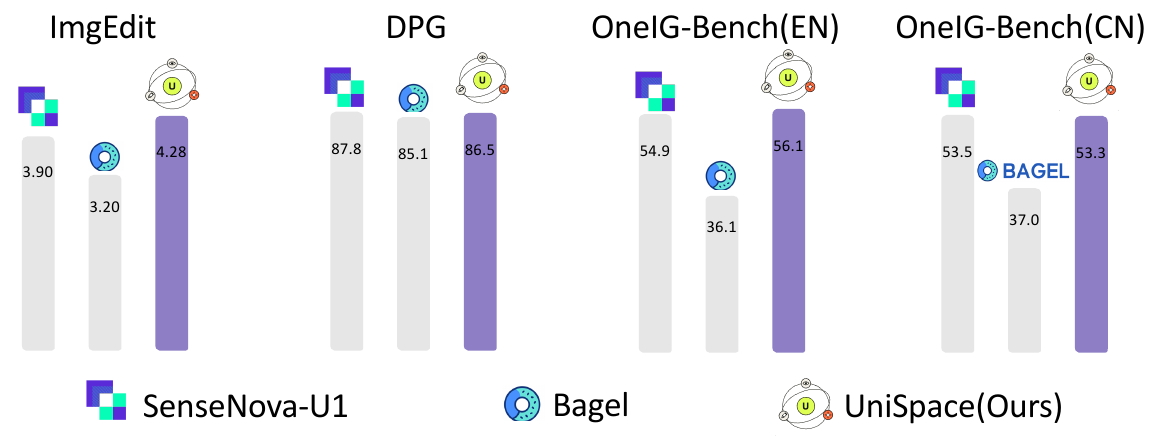}
\centering

\label{fig:teaser}
\end{figure}

\end{abstract}

\clearpage
\tableofcontents
\clearpage

\section{Introduction}
\label{sec:intro}

Semantic vision encoders have become one of the most important visual interfaces in modern multimodal systems. In vision-language models, encoders such as SigLIP~\citep{tschannen2025siglip2} and CLIP~\citep{radford2021learning} provide semantically aligned visual tokens for image understanding. Beyond understanding, they are also widely used in image generation and editing systems as semantic conditioning signals. However, these tokens are primarily optimized for semantic abstraction and alignment, rather than for preserving fine-grained visual details. Their final-layer representations often exhibit poor pixel-level recoverability, making them insufficient as the sole visual interface for detail-sensitive tasks such as image generation and reference-preserving editing. 

This limitation leads current generation and editing systems to rely on fragmented visual representations. Text-to-image models typically generate images in an autoencoding latent space, such as VAE latents, while semantic encoders are introduced separately when high-level image semantics are needed. The separation becomes more problematic for image editing, where the reference image must provide both semantic guidance, such as object identity and layout, and fine-grained details, such as texture, color, and local structure. As a result, existing systems often combine semantic encoder features with reconstruction-oriented latents or features. Recent unified multimodal models ~\citep{deng2025emerging} inherit the same issue: although understanding, generation, and editing may be placed within a single backbone, the visual information is still represented through separate semantic and reconstruction-oriented spaces. They are unified at the architecture level, but not at the level of visual representation. 

This work asks whether visual understanding, generation, and editing can be
modeled within a single visual representation space built upon a pretrained
semantic ViT. This raises two central challenges. First, we need a visual
representation that unifies semantic understanding and fine-grained detail
reconstruction while retaining the pretrained semantic backbone. Second, we
need to scale this representation to an LLM-based unified multimodal model
that supports understanding, generation, and editing in the same visual space.

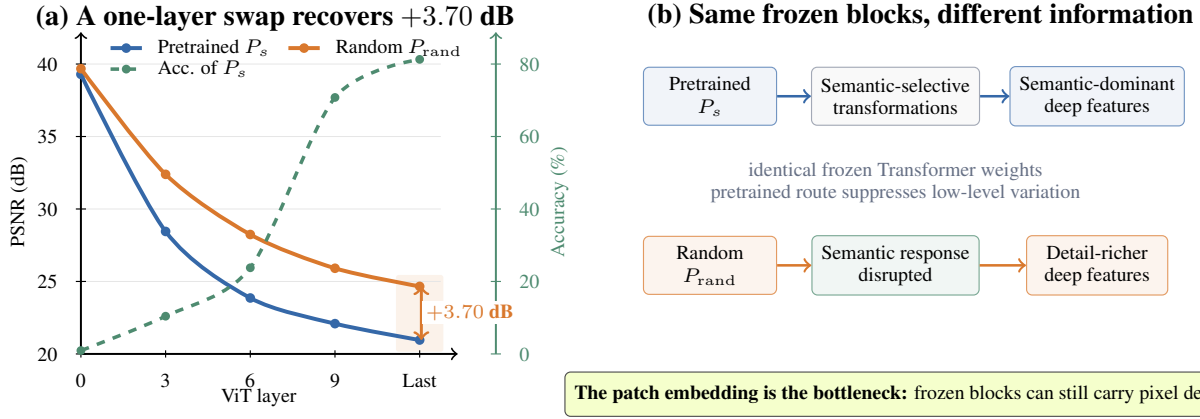
\begin{figure}[t]
\centering
\definecolor{probeBlue}{HTML}{3568A8}
\definecolor{probeOrange}{HTML}{D9782D}
\definecolor{probeGreen}{HTML}{4E8B70}
\definecolor{probeGray}{HTML}{667085}
\definecolor{probeLime}{HTML}{DFFF55}
\resizebox{\textwidth}{!}{%
\begin{tikzpicture}[font=\small]
\begin{scope}[xshift=0cm,yshift=0cm,x=1.05cm,y=0.18cm]
  \node[anchor=west,font=\bfseries] at (-0.65,23.2) {(a) A one-layer swap recovers $+3.70$ dB};

  \fill[probeOrange!10,rounded corners=1pt] (3.72,0.15) rectangle (4.28,5.45);

  \foreach \y in {0,5,10,15,20} {
    \draw[gray!20,line width=0.35pt] (-0.05,\y) -- (4.25,\y);
  }
  \draw[->,line width=0.7pt] (0,0) -- (4.45,0);
  \draw[->,line width=0.7pt] (0,0) -- (0,22.2);
  \draw[->,probeGreen,line width=0.7pt] (4.90,0) -- (4.90,22.2);

  \foreach \x/\lab in {0/0,1/3,2/6,3/9,4/Last} {
    \draw (\x,0.25) -- (\x,-0.25);
    \node[anchor=north,font=\scriptsize] at (\x,-0.65) {\lab};
  }
  \foreach \y/\lab in {0/20,5/25,10/30,15/35,20/40} {
    \draw (0.06,\y) -- (-0.06,\y);
    \node[anchor=east,font=\scriptsize] at (-0.16,\y) {\lab};
  }
  \foreach \y/\lab in {0/0,5/20,10/40,15/60,20/80} {
    \draw[probeGreen] (4.84,\y) -- (4.96,\y);
    \node[anchor=west,font=\scriptsize,probeGreen] at (5.06,\y) {\lab};
  }
  \node[font=\scriptsize] at (2.1,-3.0) {ViT layer};
  \node[rotate=90,font=\scriptsize] at (-0.73,10.5) {PSNR (dB)};
  \node[rotate=90,font=\scriptsize,probeGreen] at (5.65,10.5) {Accuracy (\%)};

  \draw[probeBlue,line width=1.4pt]
    plot[smooth] coordinates {(0,19.29) (1,8.45) (2,3.86) (3,2.09) (4,0.96)};
  \foreach \x/\y in {0/19.29,1/8.45,2/3.86,3/2.09,4/0.96}
    \fill[probeBlue] (\x,\y) circle (1.7pt);

  \draw[probeOrange,line width=1.4pt]
    plot[smooth] coordinates {(0,19.68) (1,12.39) (2,8.24) (3,5.91) (4,4.66)};
  \foreach \x/\y in {0/19.68,1/12.39,2/8.24,3/5.91,4/4.66}
    \fill[probeOrange] (\x,\y) circle (1.7pt);

  \draw[probeGreen,dashed,line width=1.25pt]
    plot[smooth] coordinates {(0,0.233) (1,2.603) (2,5.945) (3,17.694) (4,20.322)};
  \foreach \x/\y in {0/0.233,1/2.603,2/5.945,3/17.694,4/20.322}
    \fill[probeGreen] (\x,\y) circle (1.55pt);

  \draw[<->,probeOrange,line width=1.05pt] (4.02,0.96) -- (4.02,4.66);
  \node[anchor=west,font=\scriptsize\bfseries,probeOrange,fill=white,
        inner sep=1.2pt] at (4.03,2.85) {$+3.70$ dB};

  \draw[probeBlue,line width=1.4pt] (0.35,21.1) -- (0.72,21.1);
  \fill[probeBlue] (0.535,21.1) circle (1.7pt);
  \node[anchor=west,font=\scriptsize] at (0.79,21.1) {Pretrained $P_s$};
  \draw[probeOrange,line width=1.4pt] (2.45,21.1) -- (2.82,21.1);
  \fill[probeOrange] (2.635,21.1) circle (1.7pt);
  \node[anchor=west,font=\scriptsize] at (2.89,21.1) {Random $P_{\mathrm{rand}}$};
  \draw[probeGreen,dashed,line width=1.25pt] (0.35,19.65) -- (0.72,19.65);
  \fill[probeGreen] (0.535,19.65) circle (1.55pt);
  \node[anchor=west,font=\scriptsize] at (0.79,19.65) {Acc. of $P_s$};
\end{scope}

\begin{scope}[xshift=7.05cm,yshift=0.05cm]
  \node[anchor=west,font=\bfseries] at (-0.15,4.15) {(b) Same frozen blocks, different information};

  \node[draw=probeBlue!70,fill=probeBlue!7,rounded corners=2pt,
        minimum width=1.65cm,minimum height=0.72cm,align=center,font=\scriptsize]
        (ps) at (0.75,3.15) {Pretrained\\$P_s$};
  \node[draw=probeGray!60,fill=gray!5,rounded corners=2pt,
        minimum width=2.05cm,minimum height=0.72cm,align=center,font=\scriptsize]
        (sem) at (3.05,3.15) {Semantic-selective\\transformations};
  \node[draw=probeBlue!70,fill=probeBlue!7,rounded corners=2pt,
        minimum width=1.75cm,minimum height=0.72cm,align=center,font=\scriptsize]
        (deep1) at (5.55,3.15) {Semantic-dominant\\deep features};
  \draw[->,probeBlue,line width=0.9pt] (ps) -- (sem);
  \draw[->,probeBlue,line width=0.9pt] (sem) -- (deep1);
  \node[probeGray,font=\scriptsize,align=center] at (3.05,2.08)
       {identical frozen Transformer weights\\pretrained route suppresses low-level variation};

  \node[draw=probeOrange!75,fill=probeOrange!8,rounded corners=2pt,
        minimum width=1.65cm,minimum height=0.72cm,align=center,font=\scriptsize]
        (pr) at (0.75,1.05) {Random\\$P_{\mathrm{rand}}$};
  \node[draw=probeGreen!75,fill=probeGreen!8,rounded corners=2pt,
        minimum width=2.05cm,minimum height=0.72cm,align=center,font=\scriptsize]
        (res) at (3.05,1.05) {Semantic response\\disrupted};
  \node[draw=probeOrange!75,fill=probeOrange!8,rounded corners=2pt,
        minimum width=1.75cm,minimum height=0.72cm,align=center,font=\scriptsize]
        (deep2) at (5.55,1.05) {Detail-richer\\deep features};
  \draw[->,probeOrange,line width=0.9pt] (pr) -- (res);
  \draw[->,probeOrange,line width=0.9pt] (res) -- (deep2);


  \node[draw=black!70,fill=probeLime!30,rounded corners=2pt,
        minimum width=6.35cm,minimum height=0.55cm,align=center,font=\scriptsize]
        at (3.15,-0.55)
        {\textbf{The patch embedding is the bottleneck:} frozen blocks can still carry pixel detail.};
\end{scope}
\end{tikzpicture}
}
\caption{\textbf{The patch embedding---not the frozen Transformer blocks---is
the reconstruction bottleneck.} (a) As semantic accuracy rises with depth,
pixel recoverability through the pretrained SigLIP2 pathway collapses. Replacing
only its patch embedding $P_s$ with a random projection $P_{\mathrm{rand}}$
raises last-layer PSNR by $3.70$ dB; all Transformer weights remain frozen and
the reconstruction probes are identical. (b) This controlled intervention
shows that input parameterization determines which information route the same
backbone activates: the pretrained route favors semantic abstraction, whereas
disrupting it exposes a detail-richer residual pathway.}
\label{fig:patch-probe}
\end{figure}

To address the first challenge, we revisit why semantic ViTs reconstruct poorly. As shown in Fig.~\ref{fig:patch-probe}, the bottleneck lies not in the frozen Transformer blocks, but in the original semantic patch embedding, which suppresses fine-grained visual details. This motivates \emph{Patch Reparameterization}: we retain the original patch embedding for semantic understanding and add a reconstruction-aware embedding to supply visual details to the same frozen backbone. Merging the two token streams yields a unified representation for both understanding and reconstruction, which we train with a balanced flow-matching objective. Applying this design to several pretrained encoders gives PR-SigLIP2, PR-DINOv2, and PR-Qwen-ViT, which preserve semantic understanding while enabling high-fidelity reconstruction, generation, and reference-preserving image editing.

To address the second challenge, we scale PR-Qwen-ViT into UniSpace, an 8B
Qwen-based Mixture-of-Transformer-Experts multimodal model. UniSpace uses the
patch-reparameterized ViT encoder--decoder as its sole visual interface. Unlike
prior unified models that combine a semantic vision encoder with a separate VAE
latent space, UniSpace represents reference images, target images, and generated
outputs in the same unified visual representation space. This design directly
tests whether a reparameterized pretrained semantic ViT can serve as a shared
visual interface for understanding, generation, and editing at scale. Our
experiments show that UniSpace enables strong text-to-image generation and
instruction-based image editing while retaining general visual understanding
capabilities under unified generative training.

Our contributions are threefold:

\begin{itemize}
    \item We introduce Patch Reparameterization, showing that frozen pretrained semantic ViTs can preserve fine-grained visual information and can be adapted to provide both semantic and reconstruction-aware visual tokens within a single backbone.

    \item We build a family of unified visual encoders, including PR-SigLIP2, PR-DINOv2, and PR-Qwen-ViT, which provide a balanced visual representation for understanding, reconstruction, and generation.

    \item We develop UniSpace, a Qwen-based unified multimodal model powered by PR-Qwen-ViT, where a single visual encoder--decoder interface supports multimodal understanding, text-to-image generation, and instruction-based image editing.
\end{itemize}

\section{Related Work}
\label{sec:related}
\subsection{Semantic Representations for Generation and Reconstruction}

Recent works have shown that semantic representations are highly beneficial for image generation. REPA~\citep{yu2024representation} aligns intermediate diffusion-transformer features with pretrained semantic representations to accelerate convergence and improve generation quality. VA-VAE~\citep{yao2025reconstruction} introduces semantic alignment into autoencoder training, making the latent space easier for generative models to learn. RAE~\citep{zheng2025rae} further demonstrates that frozen semantic encoder features can directly serve as generative latents when paired with a learned reconstruction decoder.

Nevertheless, semantic representations are not automatically unified visual tokenizers. Such tokenizers must jointly support semantic understanding, high-fidelity reconstruction, and generative modeling. UniFlow~\citep{yue2025uniflow} constructs understanding-aware reconstruction representations through distillation, and RAEv2~\citep{singh2026improved} improves RAE-style reconstruction by incorporating multi-layer DINO ~\citep{caron2021emerging}features. These works extend semantic representations toward reconstruction, but the balance among semantic capability, reconstruction fidelity, and generative modeling remains challenging. In particular, strong understanding and reconstruction do not necessarily imply a representation that is easy for a generative prior to model. Patch Reparameterization targets this gap by improving reconstruction fidelity within a frozen semantic ViT while keeping the resulting representation suitable for generative modeling.

\subsection{Visual Representation Spaces in Unified Multimodal Models}

Recent works have explored unified multimodal models that support understanding, generation, and editing within a single framework~\citep{cui2025emu3, wang2024emu3, sensenova2026u1, deng2025emerging}. However, they differ substantially in how visual information is represented. Some methods model images with discrete visual tokenizers~\citep{cui2025emu3, wang2024emu3}, while others operate in continuous latent spaces. A common design is to use separate pretrained visual spaces for different tasks. For example, BAGEL ~\citep{deng2025emerging} adopts a Mixture-of-Transformer-Experts architecture in which semantic vision-encoder tokens support understanding and VAE latents support generation, with shared self-attention enabling interaction between the two pathways. Although the model is unified architecturally, its visual representation remains split across semantic and reconstruction-oriented spaces.

Our UniSpace instead preserves the semantic prior of a pretrained ViT and uses the patch-reparameterized encoder--decoder as the only visual tokenizer. This allows understanding, generation, and editing to operate in the same unified visual representation space while benefiting from the efficiency and transferability of pretrained semantic encoders, without introducing a separate VAE latent space.

\section{Unified Encoder via Patch Reparameterization}
\label{sec:method}

Pretrained semantic ViTs, such as SigLIP and CLIP, provide strong visual perception and image-text alignment, but their final tokens are not designed
for faithful pixel reconstruction. Existing attempts to obtain understanding-aware reconstruction representations often train a new visual
encoder~\citep{yao2025towards} or rely on semantic distillation~\citep{yue2025uniflow}, which complicates the system and can alter the original semantic representation. We ask a simpler question: is it
necessary to relearn the Transformer blocks, or can a frozen semantic ViT be
reparameterized to expose the visual details that its original semantic tokens
suppress? 


We answer this question with Patch Reparameterization. The pretrained patch
embedding is kept unchanged to preserve the original semantic pathway, while an
additional reconstruction-aware patch embedding provides an alternative
parameterization that preserves reconstruction-relevant visual information
within the same frozen Transformer blocks. Through a simple Token Fusion strategy,
where reconstruction tokens are compressed and concatenated with semantic tokens,
the resulting representation unifies semantic and reconstruction information,
providing a shared visual interface for understanding and generative modeling.

\subsection{Patch Reparameterization Can Carry Visual Details}

Prior analyzes ~\citep{singh2026improved} have suggested that shallow features in semantic ViTs preserve
more fine-grained visual details, whereas deeper features become more
semantically abstract. Consistent with this observation, Fig.~\ref{fig:patch-probe}
shows that as visual tokens propagate through deeper Transformer blocks, their
semantic classification accuracy increases from $0.93\%$ to $81.29\%$, while
their pixel-level recoverability degrades rapidly. However, this degradation does not necessarily imply that the frozen Transformer blocks are intrinsically unable to carry visual details. Unlike autoencoders with an explicit low-dimensional bottleneck, ViT backbones maintain high-dimensional token representations throughout the network. This design provides a plausible pathway for propagating input-dependent details across layers.

To test whether the poor recoverability is imposed by the frozen Transformer
blocks themselves, we conduct a diagnostic experiment on SigLIP2. We replace
its pretrained patch embedding with a randomly initialized linear projection,
keep all subsequent Transformer blocks frozen, and train identical
reconstruction probes on features extracted at different depths. If the frozen
semantic Transformer blocks were incapable of carrying pixel-level information,
changing only the input projection would not improve recoverability at deep
layers.

Counterintuitively, the random patch embedding improves the last-layer PSNR
from $20.96$ to $24.66$, despite using exactly the same frozen Transformer
blocks. As shown in Fig.~\ref{fig:patch-probe}, at the patch-embedding output, the pretrained and random projections
are similarly recoverable, achieving PSNR values of $39.29$ and $39.68$,
respectively. Their behavior diverges only as the tokens pass through the same frozen semantic Transformer blocks. Under the pretrained patch embedding, the input
activates visual patterns that the subsequent blocks were optimized to
semantically abstract, progressively suppressing variations irrelevant to the
semantic objective. The random projection disrupts this preferred semantic
processing trajectory, allowing more low-level input variation to remain
recoverable through the residual pathway.

\begin{figure}[t]
\centering
\includegraphics[width=\textwidth]{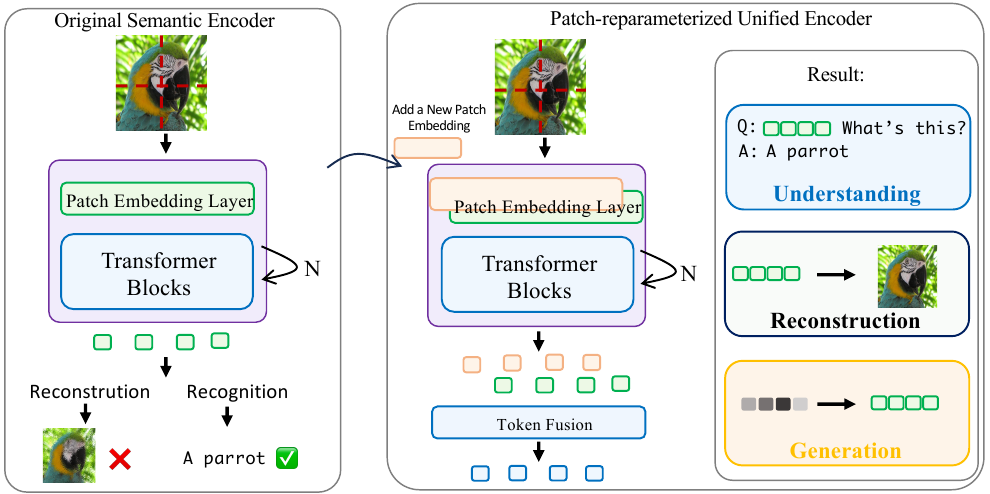}
\caption{Overview of Patch Reparameterization. The original semantic patch
embedding and frozen ViT blocks preserve the pretrained semantic pathway, while
a reconstruction-aware patch embedding injects visual details into the same
frozen backbone. The semantic and reconstruction-aware tokens are combined by a
Token Fusion layer to form the unified representation, which is
shared by understanding, reconstruction, and generation.}
\label{fig:method-overview}
\end{figure}

This result suggests that pixel-level information is not lost because the hidden space cannot carry it. Rather, it is selectively suppressed under the semantic encoder's original input
parameterization. However, randomizing the patch embedding is not itself a useful
solution, since it destroys the pretrained semantic representation. It instead
motivates a minimal adaptation: keep the original semantic patch embedding for
understanding, and learn an additional reconstruction-aware patch embedding
that injects visual details into the same frozen Transformer blocks, as shown in Fig.\ref{fig:method-overview}.

\subsection{Constructing a Unified Representation}

Given an input image $I$, we construct a unified representation that preserves
the semantic ability of the pretrained encoder while adding reconstruction
details in a channel-factorized form. Let $F_{\phi}$ denote the frozen
Transformer blocks of a pretrained semantic ViT, and let $P_s$ denote its
original pretrained patch embedding. The original semantic tokens are

\begin{equation}
\label{eq:semantic-tokens}
T_s = F_{\phi}(P_s(I)),
\end{equation}
where both $P_s$ and $F_\phi$ are kept frozen. This path preserves the
pretrained semantic representation and provides the basis for visual
understanding.

To capture reconstruction-relevant details, we introduce a reconstruction-aware
patch embedding $P_r$, initialized from $P_s$ and optimized for reconstruction.
It is followed by the same frozen Transformer blocks:
\begin{equation}
\label{eq:reconstruction-tokens}
T_r = F_{\phi}(P_r(I)).
\end{equation}
Thus, $T_r$ is not produced by an independent reconstruction encoder; it is
encoded within the same semantic ViT parameter space through a different input
parameterization.

\paragraph{Explicit Token Fusion.}
A naive way to unify $T_s$ and $T_r$ is to fuse them into a single entangled feature space using an MLP. However, as discussed in Sec.~\ref{sec:diagnostic-study}, this implicit fusion makes it difficult to control the relative contributions of semantic and reconstruction information during generative modeling. We therefore preserve the two components as explicitly separated feature groups along the channel dimension. Before concatenation, we first project the reconstruction tokens into a compact space:
\begin{equation}
\label{eq:reconstruction-projection}
\widetilde{T}_r = W_r T_r,
\end{equation}
where $W_r$ is a learnable linear projection along the channel dimension,
mapping the reconstruction tokens from $d$ to a lower dimension $d_r$. This
projection serves a distinct role from concatenation: it reduces the
dimensionality of the reconstruction component, lowering the burden for the
generative model while retaining decoder-critical visual details. We then concatenate the original semantic tokens and the compressed reconstruction tokens along the channel dimension:
\begin{equation}
\label{eq:unified-representation}
T_u = \operatorname{Concat}(T_s, \widetilde{T}_r).
\end{equation}

The complete unified representation $T_u$ serves as the shared visual
representation for all downstream tasks, including understanding,
reconstruction, and generation; neither $T_s$ nor $\widetilde{T}_r$ is used
independently. Unlike learned feature merging, this explicit fusion
leaves the pretrained semantic representation unchanged and maintains an
explicit boundary between semantic and reconstruction information. Compression
reduces the dimensionality of reconstruction details, while concatenation makes
the two information sources separately addressable during generative training.

\subsection{Training for Reconstruction and Generation}

\paragraph{Training the Reconstruction-Aware Path.}
After constructing $T_u$, we adopt a ViT decoder $D_\psi$, following RAE, to
reconstruct the input image:
\begin{equation}
\label{eq:reconstruction-decoding}
\hat{I} = D_\psi(T_u).
\end{equation}

During training, the reconstruction-aware components and the decoder are
optimized with a reconstruction objective:
\begin{equation}
\label{eq:reconstruction-loss}
\mathcal{L}_{\mathrm{rec}} = \mathcal{D}(\hat{I}, I),
\end{equation}
where $\mathcal{D}$ is instantiated with pixel-level and perceptual
reconstruction losses. Only the reconstruction-aware patch embedding $P_r$, the
linear projection $W_r$, and the decoder $D_\psi$ are updated. The original
patch embedding $P_s$ and the pretrained ViT blocks $F_\phi$ remain frozen.
Since reconstruction training never modifies the original semantic path, the
pretrained semantic tokens are structurally preserved while the additional path
learns to supply decoder-critical details.

\paragraph{Balanced Generative Modeling.} 
The explicit channel-wise decomposition of $T_u$ also gives direct control over
generative training. A plain MSE over the concatenated representation weights
the two components according to their dimensionality and scale, without
specifying how much of the objective should focus on decoder-critical
reconstruction information. We therefore use a balanced flow-matching
objective that normalizes each component by its dimension and then assigns an
explicit objective weight.

We model the distribution of the unified representation using conditional flow
matching. Let $Z_1=T_u$ denote a unified representation,
$Z_0\sim\mathcal{N}(0,\mathbf{I})$ a Gaussian noise sample of the same shape,
and $t\sim\mathcal{U}[0,1]$ a timestep. We construct the linear probability
path
\begin{equation}
\label{eq:flow-path}
Z_t=(1-t)Z_0+tZ_1,
\end{equation}
whose target velocity is
\begin{equation}
\label{eq:flow-target}
V_t=\frac{\mathrm{d}Z_t}{\mathrm{d}t}=Z_1-Z_0.
\end{equation}
Given $Z_t$, $t$, and condition $c$, the generative model $v_\theta$ predicts
the velocity $\widehat{V}_t=v_\theta(Z_t,t,c)$. Let $T_s\in\mathbb{R}^{N\times a}$ and
$\widetilde{T}_r\in\mathbb{R}^{N\times b}$ denote the semantic and
reconstruction components of $T_u$, respectively. Their explicit channel-wise
separation allows us to partition the target and predicted velocities as
\begin{equation}
V_t=\operatorname{Concat}(V_t^s,V_t^r),
\qquad
\widehat{V}_t=\operatorname{Concat}(\widehat{V}_t^s,\widehat{V}_t^r).
\end{equation}
We normalize the prediction error of each component by its number of
dimensions and optimize the reconstruction--semantic balanced flow-matching
objective
\begin{equation}
\label{eq:balanced-flow-matching}
\mathcal{L}_{\mathrm{BFM}}
=
\mathbb{E}_{Z_1,Z_0,t,c}
\left[
(1-\lambda_r)
\frac{\lVert\widehat{V}_t^s-V_t^s\rVert_F^2}{Na}
+
\lambda_r
\frac{\lVert\widehat{V}_t^r-V_t^r\rVert_F^2}{Nb}
\right],
\end{equation}
where $\lambda_r$ specifies the fraction of the total objective weight assigned
to the reconstruction component. We set $\lambda_r=0.75$, allocating $75\%$ of
the objective weight to the decoder-critical reconstruction component while
retaining the semantic component that facilitates generative learning; this
choice is supported by the ablation in Table~\ref{tab:bfm-ablation}.

At inference time, we sample $Z_0\sim\mathcal{N}(0,\mathbf{I})$ and solve the
ordinary differential equation
\begin{equation}
\frac{\mathrm{d}Z_t}{\mathrm{d}t}=v_\theta(Z_t,t,c),
\qquad t:0\rightarrow1,
\end{equation}
to obtain a generated unified representation $\widehat{T}_u$, which is decoded
into an image as $\widehat{I}=D_\psi(\widehat{T}_u)$.

\subsection{Implementation Details}

\paragraph{Unified tokenizer construction.}
We instantiate Patch Reparameterization on three pretrained semantic ViTs: SigLIP2-B~\citep{tschannen2025siglip2}, DINOv2-B~\citep{oquab2023dinov2}, and Qwen-ViT~\citep{bai2025qwen3}, obtaining PR-SigLIP2, PR-DINOv2, and PR-Qwen-ViT, respectively. SigLIP2-B is evaluated at $256\times256$ resolution with a $16\times16$ token grid and a $768$-dimensional semantic space. DINOv2-B uses its native $14\times14$ token grid with a $768$-dimensional semantic space. Qwen-ViT supports native-resolution inputs, uses a $16\times16$ patch size, and has a $1152$-dimensional semantic space in our setting. For all backbones, the reconstruction-aware patch embedding $P_r$ is initialized from the pretrained patch embedding $P_s$, while all Transformer blocks $F_\phi$ are frozen. The reconstruction projection $W_r$ maps the reconstruction-aware stream to $128$ channels, yielding a $896$ unified representation for PR-SigLIP2 and PR-DINOv2 and a $1280$ representation for PR-Qwen-ViT.

\paragraph{Reconstruction training.}
For image reconstruction, we use a ViT-XL-scale decoder with $28$ layers, hidden width $1152$, FFN width $4096$, and $16$ attention heads. Following the RAE training recipe, the reconstruction objective combines pixel-level and perceptual losses, and the decoder is further refined with an adversarial loss for high-fidelity decoding. For SigLIP2-B and DINOv2-B, we first train $P_r$, $W_r$, and the decoder using L2 and LPIPS losses. We use AdamW with learning rate $2\times10^{-4}$, $(\beta_1,\beta_2)=(0.9,0.95)$, weight decay $0$, global batch size $512$, one warmup epoch, and EMA decay $0.9978$. We then freeze the encoder-side modules and train only the decoder for another $20$ epochs with L2, LPIPS, and GAN losses, using the same optimizer, batch size, and EMA setting. The discriminator learning rate follows a cosine schedule from $2\times10^{-4}$ to $2\times10^{-5}$. The main ImageNet results additionally use a short decoder calibration stage: we sample latents from an early DiT checkpoint and update only the decoder, while keeping the encoder and generative model fixed, so that the decoder adapts to realistic generation errors in the unified latent space. SigLIP2-B and DINOv2-B are trained on ImageNet-1K at $256\times256$ resolution, while Qwen-ViT is trained on web data with the same unified-tokenizer design.

\paragraph{Generation Training}
 For ImageNet generation, we evaluate PR-SigLIP2 and PR-DINOv2 following the RAE~\citep{zheng2025rae} generation protocol, adopting the same DiTwDDTHead architecture with input size \(16\), patch size \(1\), \(896\) input channels, hidden sizes \((1152,2048)\), depths \((28,2)\), and \(16\) attention heads. We use linear velocity prediction with a logit-normal time distribution and the dimension-dependent time-shift rule from RAE~\citep{zheng2025rae}; for the $896\times16\times16$ latent, the shift uses dimension $229{,}376$ with base $4096$. The model is trained with AdamW using learning rate $2\times10^{-4}$, betas $(0.9,0.95)$, weight decay $0$, global batch size $1024$ with gradient accumulation $2$, EMA decay $0.9995$, gradient clipping $1.0$, and a linear schedule with $40$ warmup epochs decaying to $2\times10^{-5}$ by epoch $800$. Sampling uses an Euler ODE solver with $50$ steps.

\section{UniSpace: Scaling Unified Modeling in One Visual Space}
\label{sec:unispace}

Having established a representation that jointly preserves semantic and pixel-level information, we next investigate whether it can serve as the common visual interface of a large-scale multimodal model. We instantiate this idea as \textbf{UniSpace}, a unified multimodal model that performs text-to-image generation, instruction-based image editing, and image understanding in one patch-reparameterized visual representation space. Editing is especially important in this setting because it stress-tests both sides of the representation: the model must understand the reference image and instruction, perform the requested modification, and preserve fine-grained details in all irrelevant regions. UniSpace therefore uses the same frozen unified visual tokenizer for all three tasks and scales it within a Mixture-of-Transformer-Experts architecture.

\subsection{Unified MoT Architecture}

UniSpace builds on a decoder-only Qwen3-8B~\citep{yang2025qwen3} backbone and adopts the Mixture-of-Transformer-Experts (MoT) design of BAGEL~\citep{deng2025emerging}. It contains an understanding expert and a generation expert that operate on a common multimodal sequence. Token types are hard-routed to modality-specific parameters: text and conditioning-image tokens are processed by the understanding expert, whereas noised visual tokens to be predicted are processed by the generation expert. At every layer, self-attention allows tokens handled by the two experts to interact directly, retaining a bottleneck-free context across understanding and generation.

\begin{figure}[t]
\centering
\includegraphics[width=\textwidth]{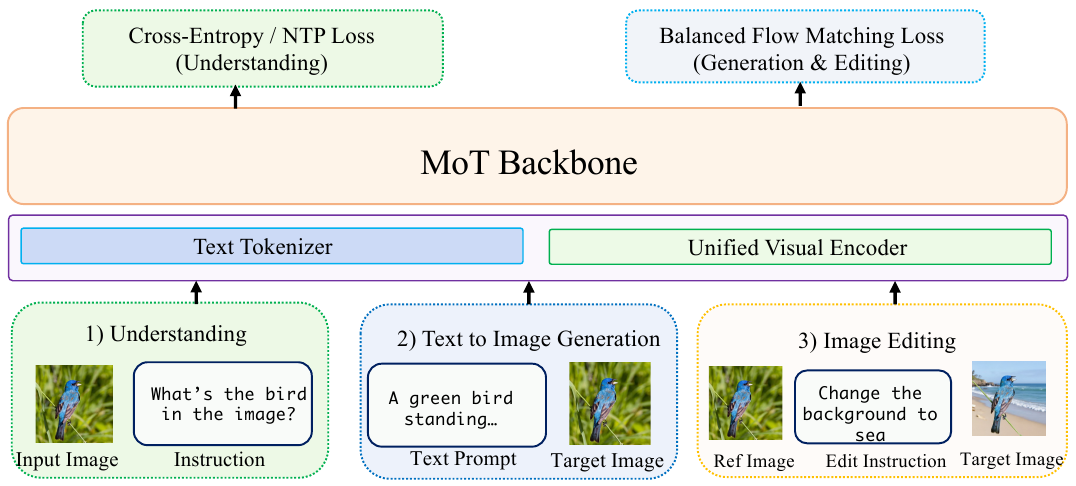}
\caption{UniSpace pipeline built on the proposed unified visual representation.
UniSpace uses the PR-Qwen-ViT encoder--decoder as the only visual
interface for reference images, target images, and generated images. Text and
visual tokens are routed through a Mixture-of-Transformer-Experts backbone for
understanding, text-to-image generation, and instruction-based image editing,
without an additional VAE pathway.}
\label{fig:unispace-architecture}
\end{figure}

The key distinction of UniSpace lies in its visual interface. Existing MoT-based models typically use two visual representation spaces: a semantic ViT supplies tokens to the understanding pathway, while a separate VAE supplies latents to the generation pathway. In contrast, UniSpace uses the unified encoder and decoder developed in Sec.~\ref{sec:method} as its only visual tokenizer, requires neither a separate VAE encoder nor an additional VAE latent space. Table~\ref{tab:unispace-visual-interface} positions this design relative to two representative MoT systems. BAGEL retains two pretrained visual spaces for semantic conditioning and generation, while SenseNova-U1 removes external visual encoders and learns a native pixel interface end-to-end. UniSpace takes a different route: it preserves the semantic prior of a pretrained ViT, reparameterizes it into a reconstruction-compatible visual tokenizer, and consolidates understanding, generation, and editing into one frozen, reusable visual representation without introducing a separate VAE pathway.

\begin{table}[t]
\centering
\caption{Comparison of visual interfaces in representative MoT-based unified
models. ``Spaces'' denotes the number of distinct visual representation spaces
used by the multimodal model.}
\label{tab:unispace-visual-interface}
\setlength{\tabcolsep}{3.2pt}
\renewcommand{\arraystretch}{1.12}
\resizebox{\textwidth}{!}{%
\begin{tabular}{lcccccc}
\toprule
Model & Visual interface & Pretrained semantic prior & Separate VAE & Spaces & Reference condition & Interface training \\
\midrule
BAGEL~\citep{deng2025emerging}
& SigLIP2 + FLUX-VAE & $\checkmark$ & $\checkmark$ & 2
& ViT + clean VAE tokens & Frozen \\
SenseNova-U1~\citep{sensenova2026u1}
& Native pixels & -- & -- & 1
& Clean pixel tokens & End-to-end \\
\textbf{UniSpace (Ours)}
& PR-Qwen-ViT tokenizer & $\checkmark$ & -- & 1
& Unified $T_u$ tokens & Frozen \\
\bottomrule
\end{tabular}%
}
\end{table}


For image understanding, an input image is encoded as $T_u$ and processed
together with the text instruction to predict the textual response. For
text-to-image generation, the target image is represented in the same $T_u$
space, corrupted along the flow path, and predicted by the generation expert
conditioned on text. Image editing combines the two cases: the reference image
is encoded as $T_u^{\mathrm{ref}}$ and provided together with the editing
instruction, while the noised target representation is predicted in the same
$T_u$ space. Through shared self-attention, the target tokens condition on both
the instruction and the reference representation, and the sampled
representation is decoded by the same decoder $D_u$. Thus, the three tasks differ in token arrangement and routing, but share the same visual tokenizer, representation space, and image decoder, as shown in Fig. \ref{fig:unispace-architecture}.

\subsection{Unified Training Objectives}

UniSpace uses two training objectives over a shared multimodal sequence: cross-entropy next-token prediction for textual responses and balanced flow matching for visual prediction in the unified representation space $T_u$. For understanding samples, the model predicts response tokens with the standard next-token prediction objective
\begin{equation}
\mathcal{L}_{\mathrm{NTP}}
=
-\sum_{k=1}^{L}
\log p_\theta(y_k \mid y_{<k}, c_{\mathrm{und}}),
\end{equation}
where $y_{1:L}$ denotes the target textual response and $c_{\mathrm{und}}$ denotes the multimodal context constructed from the input image representation $T_u$ and the text instruction. For generation and editing tasks, the target image is first mapped into the unified representation space as \(Z_1\), and then perturbed following the linear flow path in Eq.~\ref{eq:flow-path}. We sample the initial noise from a standard Gaussian distribution and define the target velocity as the displacement between the noise and target representation. Following RAE~\citep{zheng2025rae}, we adopt a logit-normal timestep sampling strategy with the same dimension-dependent time-shift rule. The model is optimized with the balanced component-wise flow matching objective in Eq.~\ref{eq:balanced-flow-matching}, where \(\lambda_r=0.75\) is used across all UniSpace training stages following the tokenizer-level generation setting.

The resulting unified visual prediction objective is

\begin{equation}
\mathcal{L}_{\mathrm{vis}}(c)
=
\mathbb{E}
\left[
\ell_{\mathrm{BFM}}
\left(
v_\theta(Z_t,t,c), V_t
\right)
\right].
\end{equation}

Here, \(c\) denotes the task-specific conditioning. For text-to-image generation, the condition contains only text tokens. For image editing, it additionally incorporates the reference image representation and the editing instruction. Thus,
\begin{equation}
\mathcal{L}_{\mathrm{t2i}}
=
\mathcal{L}_{\mathrm{vis}}(c_{\mathrm{t2i}}),
\qquad
\mathcal{L}_{\mathrm{edit}}
=
\mathcal{L}_{\mathrm{vis}}(c_{\mathrm{edit}}).
\end{equation}
Generation and editing therefore share the same target representation space and visual prediction loss, differing only in their conditioning context.

Overall, UniSpace is trained on a mixture of understanding, text-to-image generation, and image-editing samples. Let $\tau\in\{\mathrm{und},\mathrm{t2i},\mathrm{edit}\}$ denote the task type of a training sample. The total training objective can be written as
\begin{equation}
\begin{split}
\mathcal{L}_{\mathrm{UniSpace}}
&=
\mathbb{E}_{(x,\tau)\sim\mathcal{D}}
\big[
\mathbf{1}_{\tau=\mathrm{und}}\mathcal{L}_{\mathrm{NTP}}
+
\mathbf{1}_{\tau=\mathrm{t2i}}\mathcal{L}_{\mathrm{vis}}(c_{\mathrm{t2i}})
\\
&\qquad+
\mathbf{1}_{\tau=\mathrm{edit}}\mathcal{L}_{\mathrm{vis}}(c_{\mathrm{edit}})
\big].
\end{split}
\end{equation}
Thus, understanding is supervised by cross-entropy over textual responses, while generation and editing share the same balanced flow-matching objective in the unified representation space $T_u$, differing only in their conditioning context. The unified encoder and decoder remain frozen throughout UniSpace training; all MoT parameters are trainable.

visual tokenizer remains frozen throughout.

\subsection{Training Details}
\label{sec:training-details}

\paragraph{Training Data and Curriculum.}
We train UniSpace on internally curated datasets for text-to-image generation,
image editing, and visual understanding. These data provide supervision for
image generation, instruction-guided editing, and image-text understanding,
respectively. The complete training curriculum is summarized in
Table~\ref{tab:unispace-training}. We progressively scale training across
three stages with increasing image resolutions of $256$, $512$, and
$1024$.

In Stage~1, we jointly optimize text-to-image generation and visual
understanding with a sampling ratio of $10{:}1$. In Stage~2, we introduce
image editing and mix generation, editing, and understanding data with a
sampling ratio of $10{:}3{:}1$. Stage~3 further scales all three capabilities
through high-resolution packed multimodal training at $1024$ resolution. Following pretraining, we perform a multimodal supervised fine-tuning (SFT)
stage using packed text-to-image, image-editing, and LLaVA-NeXT-style
visual instruction data~\citep{liu2024llavanext}. During SFT, generation,
editing, and visual instruction samples are mixed with a sampling ratio of
$10{:}3{:}2$. This stage improves instruction following and visual
understanding while preserving the generation and editing capabilities
acquired during pretraining.

Across all training stages, UniSpace processes approximately $510$M training
sample instances, corresponding to approximately $470$B multimodal tokens.
Here, sample instances refer to examples observed during training and may
include repeated draws from the underlying datasets. The unified visual
tokenizer and visual encoder remain frozen throughout all stages.

\begin{table*}[t]
\centering
\caption{Progressive training curriculum of UniSpace. T2I, Edit, Und., and VLM
denote text-to-image generation, image editing, visual understanding, and visual
instruction tuning, respectively. Sample instances refer to the number of
examples processed during training rather than the number of unique examples.}
\label{tab:unispace-training}
\setlength{\tabcolsep}{7.0pt}
\renewcommand{\arraystretch}{1.14}
\begin{tabular}{lcccc}
\toprule
\textbf{Item} & \textbf{Stage 1} & \textbf{Stage 2} & \textbf{Stage 3} & \textbf{SFT} \\
\midrule
Resolution
& $256$
& $512$
& $1024$
& $1024$ \\

Training tasks
& T2I + Und.
& T2I + Edit + Und.
& T2I + Edit + Und.
& T2I + Edit + VLM \\

Sampling ratio
& $10{:}1$
& $10{:}3{:}1$
& $10{:}3{:}1$
& $10{:}3{:}2$ \\

Optimization steps
& 170K
& 115K
& 60K
& 12K \\

Sample instances
& 264.2M
& 189.9M
& 50.9M
& 4.9M \\

Training tokens
& 110B
& 196B
& 151B
& 15B \\

Peak learning rate
& $2\times10^{-4}$
& $1\times10^{-4}$
& $4\times10^{-5}$
& $2\times10^{-5}$ \\

Maximum sequence length
& 3,072
& 8,192
& 12,288
& 13,000 \\

Hardware
& 256 NPUs
& 256 NPUs
& 256 NPUs
& 128 NPUs \\
\bottomrule
\end{tabular}
\end{table*}

\paragraph{Optimization.}
We optimize UniSpace using AdamW with $\beta_1=0.9$, $\beta_2=0.95$,
$\epsilon=10^{-15}$, and zero weight decay. We use a linear warmup followed
by a constant learning-rate schedule. The warmup length is 600 steps for the
three pretraining stages and 100 steps for SFT. The peak learning rates for
each stage are reported in Table~\ref{tab:unispace-training}. We train in
BF16 precision and clip the global gradient norm to $1.0$. We use one packed
sequence per device at each optimization step without gradient accumulation.
No exponential moving average is used during training.

The model is trained progressively: each stage is initialized from the selected
checkpoint of the preceding stage, while the optimizer and learning-rate
scheduler are reinitialized at the beginning of each new stage. The visual
encoder and the unified visual tokenizer are frozen throughout training, while
the language model, modality experts, and multimodal projection modules are
optimized jointly.

\paragraph{Packed Multimodal Training.}
To improve training efficiency under heterogeneous multimodal sequence lengths,
we employ packed multimodal training. Multiple independent examples, potentially
from different training objectives, are concatenated into a single packed
sequence until a predefined token budget is reached. Samples are drawn according
to the task-specific sampling ratios in Table~\ref{tab:unispace-training}.
Consequently, the specified ratios describe the expected sampling frequencies
over the full training run rather than the exact composition of every individual
packed sequence.

We construct block-wise attention masks for packed sequences to prevent
information exchange between different examples. In particular, tokens may only
attend to tokens belonging to the same example, while attention within each
example follows its corresponding modality-specific pattern. Text tokens use
causal attention, whereas visual tokens use full attention within the same
sample. Position indices are reset for every individual example inside a packed
sequence. Losses are computed only over valid target tokens: the text
cross-entropy loss is applied to target text tokens, while the image generation
loss is applied only to target visual tokens.

The maximum packed sequence lengths are $3{,}072$, $8{,}192$,
$12{,}288$, and $13{,}000$ for Stage~1, Stage~2, Stage~3, and SFT,
respectively. The resulting token utilization is approximately $73\%$ to
$79\%$ across stages. As the image resolution increases, each packed sequence
contains fewer examples due to the increased number of visual tokens; however,
the global task sampling distribution remains unchanged.

\paragraph{Data Processing.}
We use resolution-specific image buckets for the three pretraining stages.
Images are resized and bucketed according to their aspect ratios, with an
aspect-ratio deviation threshold of $0.05$. Visual inputs are tokenized using
the frozen unified visual tokenizer. For image generation, we apply text
conditioning dropout with probability $0.1$ to support classifier-free
guidance. We also apply instruction dropout with probability $0.1$ during
multimodal training. For image editing, reference images are processed through
the understanding branch, while the target image is represented as a generation
target.

\paragraph{Distributed Training Infrastructure.}
We train Stages~1--3 on 256 Ascend 910B NPUs with 64GB memory each, distributed
over 16 nodes with 16 NPUs per node. The SFT stage is trained on 128 Ascend
910B NPUs over 8 nodes. Training is implemented in PyTorch using FSDP with
the \texttt{SHARD\_GRAD\_OP} sharding strategy. We use BF16 mixed precision,
activation checkpointing, and SDPA-based attention to reduce memory consumption.
The selected training trajectory requires approximately 142K NPU-hours in
total. 
\section{Experiments}
\label{sec:exp}

\subsection{Unified Tokenizer Evaluation}
For reconstruction evaluation, we use the ImageNet-1K validation set to assess PR-SigLIP2 and PR-DINOv2, reporting PSNR, rFID, and LPIPS as evaluation metrics.  For multimodal understanding, we follow the LLaVA-v1.5~\citep{liu2024improved} setup, using Vicuna-7B-v1.5 as the language model and a two-layer MLP as the multimodal projector. The visual encoder is frozen, and the complete unified representation $T_u$ is provided as input to the projector.  For ImageNet generation, we evaluate our method on ImageNet using rFID. We report results from checkpoints trained for 80 and 800 epochs.
\subsubsection{High-Fidelity Image Reconstruction}


\begin{table}[t]
\centering
\caption{Reconstruction quality on the ImageNet-1K $256\times256$ validation set. ``Ratio'' denotes the spatial downsampling ratio.}
\label{tab:imagenet-reconstruction}
\setlength{\tabcolsep}{3.2pt}
\renewcommand{\arraystretch}{1.12}
\resizebox{\textwidth}{!}{%
\begin{tabular}{lcccrrr}
\toprule
Method & Type & Training Data & Ratio & PSNR$\uparrow$ & SSIM$\uparrow$ & rFID$\downarrow$ \\
\midrule
\multicolumn{7}{c}{\emph{Tokenizers without Demonstrated Semantic Capability}} \\
\midrule
LlamaGen~\citep{sun2024autoregressive} & Discrete-Pixel & MS+IN-1K & 16 & 20.65 & 0.54 & 2.47 \\
Open-MAGVIT2~\citep{luo2024open} & Discrete-Pixel & Mixed-100M & 16 & 22.70 & 0.64 & 1.67 \\
SD-VAE XL~\citep{rombach2021highresolution} & Continuous-Pixel & OImg+LAae++ & 8 & 27.37 & 0.78 & 0.67 \\
Qwen-Image~\citep{wu2025qwen} & Continuous-Pixel & -- & 8 & \textbf{32.18} & 0.90 & 1.45 \\
SD-VAE 3~\citep{rombach2021highresolution} & Continuous-Pixel & -- & 8 & 31.29 & 0.87 & 0.20 \\
Wan2.1~\citep{wan2025} & Continuous-Pixel & -- & 8 & 31.34 & 0.89 & 0.95 \\
FLUX-VAE~\citep{flux2024} & Continuous-Pixel & -- & 8 & 32.74 & \textbf{0.92} & \textbf{0.18} \\
VA-VAE~\citep{yao2025reconstruction} & Continuous-Pixel & IN-1K & 16 & 27.96 & 0.79 & 0.28 \\
Wan2.2~\citep{wan2025} & Continuous-Pixel & -- & 16 & 31.25 & 0.88 & 0.74 \\
\midrule
\multicolumn{7}{c}{\emph{Tokenizers with Demonstrated Semantic Capability}} \\
\midrule

Show-o~\citep{xie2025show} & Discrete-Pixel & -- & 16 & 21.34 & 0.59 & 3.50 \\
QLIP-B~\citep{zhao2025qlip} & Discrete-Pixel & DC-1B & 16 & 23.16 & 0.63 & 3.21 \\
VILA-U~\citep{wu2025vila} & Discrete-Pixel & WL-10B+CY-1B & 16 & -- & -- & 1.80 \\
TokenFlow~\citep{qu2025tokenflow} & Discrete-Pixel & LA+CY & 16 & 21.41 & 0.69 & 1.37 \\
UniTok~\citep{ma2026unitok} & Discrete-Pixel & DC-1B & 16 & 27.28 & 0.77 & 0.41 \\
UniLIP~\citep{tang2025unilip} & Continuous-Pixel & BP-32M & 32 & 22.99 & 0.75 & 0.79 \\
BLIP3-o~\citep{chen2025blip3} & Continuous-Diffusion & BP-32M & 16 & 14.71 & 0.58 & 3.18 \\
VTP-L~\citep{yao2025towards} & Continuous-Pixel & DC-277M & 16 & 25.82 & 0.74 & 0.36 \\
UniFlow (SigLIP2)~\citep{yue2025uniflow} & Continuous-Diffusion & IN-1K & 16 & 29.38 & 0.93 & 0.62 \\
UniFlow (DINOv2)~\citep{yue2025uniflow} & Continuous-Diffusion & IN-1K & 14 & \textbf{31.01} & \textbf{0.94} & 0.54 \\
RAE (SigLIP2-B)~\citep{zheng2025rae} & Continuous-Pixel & IN-1K & 16 & 19.35 & 0.49 & 0.53 \\
RAE (DINOv2-B)~\citep{zheng2025rae} & Continuous-Pixel & IN-1K & 14 & 18.86 & 0.48 & 0.57 \\
RAEv2 (DINOv3-L, $K{=}7$)$^\dagger$~\citep{singh2026improved} & Continuous-Pixel & IN-1K & 16 & 22.57 & 0.63 & 0.29 \\
\textbf{PR-SigLIP2} & Continuous-Pixel & IN-1K & 16 & 29.64 & 0.87 & 0.18 \\
\textbf{PR-DINOv2} & Continuous-Pixel & IN-1K & 14 & 30.84 & 0.90 & \textbf{0.14} \\
\textbf{PR-Qwen-ViT} & Continuous-Pixel & Web Data & 16 & 30.16 & 0.88 & 0.17 \\
\bottomrule
\end{tabular}%
}
\vspace{2pt}
\begin{minipage}{0.98\textwidth}
\footnotesize
\emph{Note.} $^\dagger$ For RAEv2, we report the $K{=}7$ variant, which is used as its reconstruction--generation trade-off setting. VTP-L reports semantic capability through ImageNet zero-shot and linear probing evaluation.
\end{minipage}
\end{table}

\begin{figure}[p]
\centering
\includegraphics[width=\linewidth]{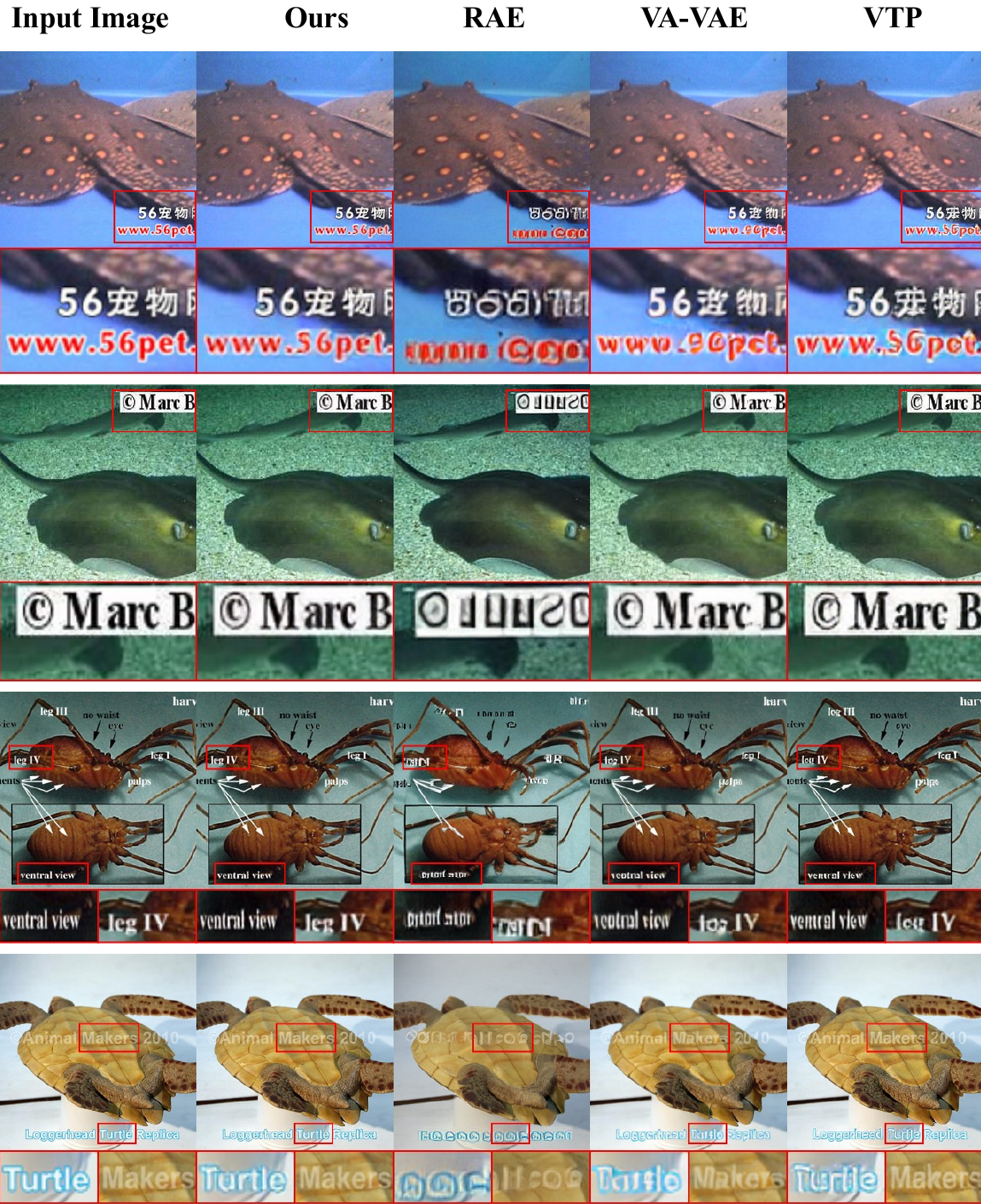}
\caption{Qualitative reconstruction comparison. Columns show the input image,
Ours, RAE, VA-VAE, and VTP. Red boxes highlight local details and text-like
regions.}
\label{fig:reconstruction-qual}
\end{figure}

We first assess whether Patch Reparameterization equips pretrained semantic
encoders with high-fidelity image reconstruction capability. As shown in
Table~\ref{tab:imagenet-reconstruction}, the proposed method consistently
achieves strong reconstruction quality across PR-SigLIP2, PR-DINOv2, and
PR-Qwen-ViT. In particular, PR-DINOv2 attains an rFID of $0.14$, a PSNR of
$30.84$, and an SSIM of $0.90$, achieving the best rFID among the compared
tokenizers with demonstrated semantic capability. Figure~\ref{fig:reconstruction-qual}
presents qualitative comparisons on representative ImageNet validation images,
with red boxes highlighting fine local details and text-like regions that are
particularly sensitive to reconstruction quality.

The matched-backbone comparisons further demonstrate the effectiveness of Patch
Reparameterization. Compared with RAE using the corresponding SigLIP2-B and
DINOv2-B encoders, PR-SigLIP2 reduces rFID from $0.53$ to $0.18$, while
PR-DINOv2 reduces it from $0.57$ to $0.14$, corresponding to relative
reductions of $66.0\%$ and $75.4\%$, respectively. For PR-DINOv2, PSNR
increases from $18.86$ to $30.84$ and SSIM from $0.48$ to $0.90$. Our method also compares favorably with the recent RAEv2
reconstruction-enhanced variant: PR-DINOv2 improves PSNR from $22.57$ to
$30.84$ and rFID from $0.29$ to $0.14$, while using the smaller DINOv2-B
encoder rather than the DINOv3-L backbone used by RAEv2. These improvements are
achieved with the pretrained ViT backbone kept frozen, indicating that its
Transformer blocks can effectively propagate reconstruction-relevant visual
details when driven by a reconstruction-aware patch embedding.

The proposed representations are also competitive with tokenizers specifically
designed for pixel reconstruction and generation. PR-DINOv2 improves over
VA-VAE by reducing rFID from $0.28$ to $0.14$, and even surpasses
large-scale generative VAEs such as FLUX-VAE and SD-VAE 3 in terms of rFID.
Meanwhile, PR-Qwen-ViT shows that Patch Reparameterization generalizes beyond
ImageNet-pretrained encoders. Crucially, unlike reconstruction-specialized
tokenizers, our representations retain the semantic capabilities of their
respective pretrained encoders, as evaluated in the next section.

\subsubsection{Multimodal Understanding}

We next examine whether Patch Reparameterization compromises the semantic understanding inherited from the pretrained encoder. We compare against the corresponding original semantic encoders under the same LLaVA-v1.5 setup, using identical language models, multimodal projectors, training data, and evaluation protocols. In all experiments, the complete unified representation $T_u$, rather than the semantic component $T_s$ alone, is fed to the multimodal projector.

As shown in Table~\ref{tab:multimodal-understanding}, Patch Reparameterization does not compromise the overall multimodal understanding capability of the original encoders. PR-SigLIP2 achieves an average score of $64.37$, surpassing the original SigLIP2-B baseline of $63.39$, while PR-Qwen-ViT reaches $68.94$, slightly above the original Qwen-ViT baseline of $68.29$. Beyond the matched-backbone comparisons, our method also achieves highly competitive understanding performance under the LLaVA-v1.5/Vicuna-7B setting. These results demonstrate that Patch Reparameterization retains the semantic capability of pretrained encoders while incorporating reconstruction-aware visual details into the unified representation.

\begin{table}[t]
\centering
\caption{Multimodal understanding performance of unified visual tokenizers. Our patch-reparameterized tokenizers are evaluated using the complete unified representation \(T_u\). Methods marked with \(\dagger\) use the LLaVA-v1.5~\citep{liu2024improved} training data. The ``Original Semantic Encoders'' rows are matched baselines using the same semantic encoder, LLM, training data, and evaluation protocol as our corresponding variants.}
\label{tab:multimodal-understanding}
\setlength{\tabcolsep}{2.2pt}
\renewcommand{\arraystretch}{1.20}
\resizebox{\textwidth}{!}{%
\begin{tabular}{lccrrrrrrrrr}
\toprule
Method & Visual Encoder & LLM & Res. & POPE & GQA & TQA & MMV & MMB & MME-S & MME-P & Avg. \\
\midrule
\multicolumn{12}{c}{\emph{Existing Unified Visual Tokenizers}} \\
\midrule
VILA-U $\dagger$~\citep{wu2025vila} & SigLIP-SO400M & Vicuna-7B & 256 & 81.6 & -- & -- & -- & -- & -- & 1311.6 & -- \\
UniTok $\dagger$~\citep{ma2026unitok} & ViTamin-L & Vicuna-7B & 256 & 81.7 & -- & -- & -- & -- & -- & 1448.0 & -- \\
QLIP $\dagger$~\citep{zhao2025qlip} & CLIP-L & Vicuna-7B & 392 & 86.1 & 61.8 & 55.2 & 33.3 & -- & -- & 1498.3 & -- \\
TokenFlow-B $\dagger$~\citep{qu2025tokenflow} & CLIP-B & Vicuna-13B & 224 & 84.0 & 59.3 & 49.8 & 22.4 & 55.3 & 1660.4 & 1353.6 & 60.21 \\
TokenFlow-L $\dagger$~\citep{qu2025tokenflow} & ViTamin-XL & Vicuna-13B & 256 & 85.0 & 60.3 & 54.1 & 27.7 & 60.3 & 1622.9 & 1365.4 & 62.40 \\
UniTok~\citep{ma2026unitok} & ViTamin-L & LLaMA-2-7B & 256 & 83.2 & 61.1 & 51.6 & 33.9 & -- & -- & 1448.0 & -- \\
TokLIP~\citep{lin2025toklip} & VQ-GAN+ViT-SO400M & Qwen2.5-7B & 384 & 84.1 & 59.5 & -- & 29.8 & 67.6 & -- & 1448.4 & -- \\
TokenFlow-XL~\citep{qu2025tokenflow} & SigLIP-SO400M & Qwen2.5-14B & 384 & 87.8 & 62.5 & 62.3 & 48.2 & 76.8 & 1922.2 & 1551.1 & 73.04 \\
UniFlow-LV $\dagger$~\citep{yue2025uniflow} & DFN-CLIP-L & Vicuna-7B & 224 & 86.56 & 61.38 & 53.40 & 30.2 & 63.83 & 1748.0 & 1446.9 & 65.02 \\
UniFlow-LV $\dagger$~\citep{yue2025uniflow} & SigLIP2-SO400M & Vicuna-7B & 256 & 87.94 & 63.29 & 58.0 & 32.4 & 68.38 & 1823.0 & 1477.9 & 67.87 \\
UniFlow-LV $\dagger$~\citep{yue2025uniflow} & DINOv2-L & Vicuna-7B & 378 & 88.04 & 59.37 & 45.53 & 25.6 & 51.48 & 1590.5 & 1257.7 & 58.92 \\
UniFlow-LV $\dagger$~\citep{yue2025uniflow} & InternViT-300M~\citep{chen2024internvl} & Vicuna-7B & 448 & 88.97 & 63.35 & 61.85 & 36.6 & 67.10 & 1803.0 & 1505.1 & 69.04 \\
\midrule
\multicolumn{12}{c}{\emph{Original Semantic Encoders}} \\
\midrule
SigLIP2 baseline $\dagger$ & SigLIP2-B & Vicuna-7B & 256 & 85.2 & 61.4 & 54.36 & 25.0 & 64.4 & 1689.3 & 1378.2 & 63.39 \\
Qwen-ViT baseline $\dagger$ & Qwen-ViT & Vicuna-7B & 448 & 86.26 & 63.25 & 64.27 & 31.88 & 69.24 & 1791.3 & 1471.3 & 68.29 \\
\midrule
\multicolumn{12}{c}{\emph{Ours: Patch-Reparameterized Unified Visual Tokenizers}} \\
\midrule
\textbf{PR-SigLIP2} $\dagger$ & PR-SigLIP2 & Vicuna-7B & 256 & 84.8 & 61.1 & 54.6 & 25.5 & 65.2 & 1741.1 & 1447.2 & 64.37 \\
\textbf{PR-Qwen-ViT} $\dagger$ & PR-Qwen-ViT & Vicuna-7B & 448 & 86.57 & 63.13 & 63.68 & 30.28 & 69.07 & 1874.5 & 1522.7 & 68.94 \\
\bottomrule
\end{tabular}%
}
\vspace{2pt}

\begin{minipage}{\textwidth}
\footnotesize
\textit{Note:} The average is computed as
$\mathrm{Avg}=(\mathrm{POPE}+\mathrm{GQA}+\mathrm{TQA}+\mathrm{MMV}+\mathrm{MMB}+\mathrm{MME\text{-}S}/20+\mathrm{MME\text{-}P}/20)/7$.
\end{minipage}
\end{table}

\subsubsection{ImageNet Generation}

\begin{table}[t]
\centering
\caption{System-level reconstruction--generation performance on ImageNet $256\times256$ with representative latent generative priors, following the evaluation protocols reported by VA-VAE~\citep{yao2025reconstruction} and RAE~\citep{zheng2025rae}.}
\label{tab:imagenet-system-performance}
\setlength{\tabcolsep}{2.4pt}
\renewcommand{\arraystretch}{1.15}
\resizebox{\textwidth}{!}{%
\begin{tabular}{lccccccccccccc}
\toprule
& & \multicolumn{1}{c}{Reconstruction} & \multicolumn{1}{c}{Model} & \multicolumn{5}{c}{Generation w/o CFG} & \multicolumn{5}{c}{Generation w/ CFG} \\
\cmidrule(lr){3-3}\cmidrule(lr){4-4}\cmidrule(lr){5-9}\cmidrule(lr){10-14}
Method & Tokenizer & rFID$\downarrow$ & \#Params & gFID$\downarrow$ & sFID$\downarrow$ & IS$\uparrow$ & Pre.$\uparrow$ & Rec.$\uparrow$ & gFID$\downarrow$ & sFID$\downarrow$ & IS$\uparrow$ & Pre.$\uparrow$ & Rec.$\uparrow$ \\
\midrule
\multicolumn{14}{c}{\emph{Representative Latent Generative Priors}} \\
\midrule
REPA~\citep{yu2024representation} & SD-VAE & 0.61 & 675M & 5.90 & -- & -- & -- & -- & 1.42 & 4.70 & 305.7 & 0.80 & 0.65 \\
LightningDiT~\citep{yao2025reconstruction} & VA-VAE & 0.28 & 675M & 2.17 & 4.36 & 205.6 & 0.77 & 0.65 & 1.35 & 4.15 & 295.3 & 0.79 & 0.65 \\
UniFlow (MAR)~\citep{yue2025uniflow} & UniFlow (InternViT) & 0.28 & 479M & 2.45 & -- & 228.0 & -- & -- & 1.85 & -- & 290.0 & -- & -- \\
VTP-L (LightningDiT)~\citep{yao2025towards} & VTP-L & 0.36 & 675M & 1.85 & -- & 232.3 & 0.79 & 0.63 & 1.11 & -- & 279.5 & 0.79 & 0.67 \\
RAE (DiT$^{\mathrm{DH}}$-XL)~\citep{zheng2025rae} & RAE (DINOv2-B) & 0.57 & 839M & 1.51 & -- & 242.9 & 0.79 & 0.63 & 1.13 & -- & 262.6 & 0.78 & 0.67 \\
RAEv2 (DiT$^{\mathrm{DH}}$-XL)~\citep{singh2026improved} & RAEv2 (DINOv3-L, $K{=}7$) & 0.29 & 0.9B & 1.65 & -- & 228.0 & -- & -- & 1.06 & -- & 255.3 & -- & -- \\
\midrule
\multicolumn{14}{c}{\emph{Ours: Patch-Reparameterized Unified Visual Tokenizers}} \\
\midrule
\textbf{PR-SigLIP2} & PR-SigLIP2 & 0.18 & 839M & 4.42 & 6.66 & 190.4 & 0.72 & 0.65 & 2.80 & 6.05 & 248.2 & 0.78 & 0.61 \\
\textbf{PR-DINOv2} & PR-DINOv2 & 0.14 & 839M & 2.10 & 5.39 & 217.2 & 0.78 & 0.64 & 1.87 & 4.89 & 274.3 & 0.82 & 0.60 \\
\bottomrule
\end{tabular}%
}
\vspace{2pt}
\begin{minipage}{0.98\textwidth}
\footnotesize
\emph{Note.} For RAEv2, we report the $K{=}7$ reconstruction--generation trade-off setting from the original paper. ``--'' indicates metrics not reported in the corresponding source under the matched setting.
\end{minipage}
\end{table}



More importantly, our method establishes a favorable
reconstruction--generation operating point for latent generative modeling. As
shown in Table~\ref{tab:imagenet-system-performance}, PR-DINOv2 shifts the
trade-off toward substantially higher reconstruction fidelity while retaining
effective generation performance, achieving an rFID of $0.14$ and gFIDs of
$2.10$ without classifier-free guidance and $1.87$ with classifier-free
guidance at scale $1.2$. In contrast, RAE, RAEv2, and VTP-L achieve stronger
generation results but have substantially higher tokenizer rFID values of
$0.57$, $0.29$, and $0.36$, respectively. Thus, Patch Reparameterization
recovers substantially more decoder-relevant visual detail from a pretrained
semantic ViT while still producing a latent space that can be effectively
modeled by a generative prior.

Taken together with the reconstruction and multimodal understanding results,
these findings demonstrate that the proposed representation is not merely a
reconstruction tokenizer or a generation latent. Instead, it provides a
unified visual representation that retains multimodal understanding,
supports high-fidelity reconstruction, and remains effective for latent image
generation.

\subsubsection{Auxiliary Diagnostic Study on Entangled Unified Representations}
\label{sec:diagnostic-study}

\begin{figure}[t]
\centering
\includegraphics[width=\textwidth]{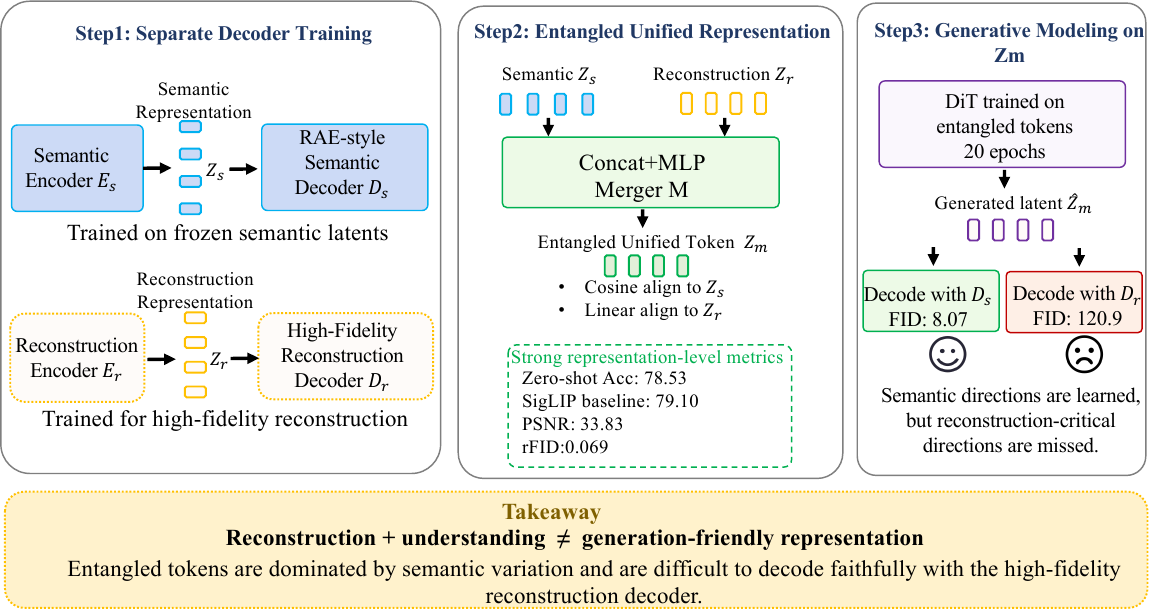}
\caption{Diagnostic study of an entangled unified representation. We construct
an MLP-merged token $Z_m=M(Z_s,Z_r)$ from semantic and reconstruction
representations and align it with both. Although $Z_m$ supports strong
understanding and high-fidelity reconstruction on real encoded latents, its
generative decoding fails when using the pretrained high-fidelity
reconstruction decoder $D_r$, which yields an FID of $120.9$. Decoding the same
generated latent with the semantic decoder $D_s$ gives a much lower FID of
$8.07$, indicating that the generative prior captures semantic structure but
fails to preserve reconstruction-relevant information in the entangled latent
space.}
\label{fig:diagnostic-entangled}
\end{figure}

As an auxiliary diagnostic experiment, separate from our final design, we
investigate whether jointly supporting semantic understanding and
high-fidelity reconstruction is alone sufficient for generative modeling. We
construct a deliberately entangled representation by merging semantic and
reconstruction features into a single latent space without explicit
decomposition. As shown in Fig.~\ref{fig:diagnostic-entangled}, this diagnostic
variant performs well when evaluated on real encoded latents, achieving a
zero-shot accuracy of $78.53$ (compared with $79.10$ for the SigLIP baseline),
an rFID of $0.069$, and a PSNR of $33.83$.

However, the entangled representation fails to support faithful generative
decoding. After training a DiT prior on the merged representation, decoding
the generated latent $\widehat{Z}_m$ with the pretrained high-fidelity
reconstruction decoder $D_r$ results in an FID of $120.9$, indicating a severe
generation failure. Interestingly, decoding the same $\widehat{Z}_m$ with the
semantic decoder $D_s$ yields a much lower FID of $8.07$. This result does not
indicate successful generation in the unified latent space; rather, it shows
that the generative prior mainly captures the dominant semantic structure,
while failing to preserve the reconstruction-relevant information required by
$D_r$.

We further find that approximately $95\%$ of the merger output variation is
explained by the semantic pathway. This semantic dominance explains the
decoder-dependent generation gap: although the generated latent retains enough
semantic structure for $D_s$ to produce plausible images, it does not preserve
the fine-grained directions needed for high-fidelity decoding by $D_r$. These
results reveal a gap between representation quality and generative
modelability: good reconstruction on real encoded latents does not guarantee
faithful reconstruction after generation. This motivates our explicit
factorization, in which $T_s$ and $\widetilde{T}_r$ are concatenated along the
channel dimension, making the reconstruction component directly accessible
during flow-matching training. Further details of the entangled representation
and diagnostic protocol are provided in Appendix~\ref{app:entangled-diagnostic}.

\begin{table}[t]
\centering
\caption{Quantitative comparison on ImgEdit. All category scores and the
overall score are higher-is-better. Baseline values are reproduced from the
evaluation reported by SenseNova-U1~\citep{sensenova2026u1}. ``Active'' denotes activated generation
parameters for sparse models.}
\label{tab:imgedit}
\setlength{\tabcolsep}{2.0pt}
\renewcommand{\arraystretch}{1.08}
\scriptsize
\resizebox{\textwidth}{!}{%
\begin{tabular}{lcccccccccccc}
\toprule
Model & \# Params & Add & Adjust & Extract & Replace & Remove & Bg. & Style & Hybrid & Action & Overall$\uparrow$ \\
\midrule
\multicolumn{12}{c}{\emph{Closed-source Models}} \\
\midrule
UniWorld-V2~\citep{li2025uniworld}              & -- & 4.29 & 4.44 & \textbf{4.32} & \textbf{4.69} & \textbf{4.72} & \underline{4.41} & 4.91 & \underline{3.83} & \underline{4.83} & \textbf{4.49} \\
Nano-Banana-Pro          & -- & 4.44 & \underline{4.62} & 3.42 & 4.60 & \underline{4.63} & 4.32 & \textbf{4.97} & 3.64 & 4.69 & \underline{4.37} \\
Seedream 4.5             & -- & 4.57 & \textbf{4.65} & 2.97 & \underline{4.66} & 4.46 & 4.37 & 4.92 & 3.71 & 4.56 & 4.32 \\
Seedream 4.0             & -- & 4.33 & 4.38 & 3.89 & 4.65 & 4.57 & 4.35 & 4.22 & 3.71 & 4.61 & 4.30 \\
Nano-Banana              & -- & \textbf{4.62} & 4.41 & 3.68 & 4.34 & 4.39 & 4.40 & 4.18 & 3.72 & \underline{4.83} & 4.29 \\
GPT-Image-1              & -- & \underline{4.61} & 4.33 & 2.90 & 4.35 & 3.66 & \textbf{4.57} & \underline{4.93} & \textbf{3.96} & \textbf{4.89} & 4.20 \\
FLUX.1 Kontext [Pro]~\citep{labs2025flux1kontextflowmatching}     & -- & 4.25 & 4.15 & 2.35 & 4.56 & 3.57 & 4.26 & 4.57 & 3.68 & 4.63 & 4.00 \\
\midrule
\multicolumn{12}{c}{\emph{Open-source Image Generation and Editing Models}} \\
\midrule
Qwen-Image-Edit-2511~\citep{wu2025qwen}     & 20B       & \textbf{4.54} & \textbf{4.57} & \underline{4.13} & \underline{4.70} & 4.46 & 4.36 & \underline{4.89} & \textbf{4.16} & \underline{4.81} & \textbf{4.51} \\
LongCat-Image-Edit~\citep{team2025longcat}       & 6B        & 4.44 & \underline{4.53} & 3.83 & \textbf{4.80} & \underline{4.60} & 4.33 & \textbf{4.92} & 3.75 & \textbf{4.82} & \underline{4.45} \\
FLUX.2 [Dev]~\citep{flux2024}             & 32B       & \underline{4.50} & 4.18 & 3.83 & 4.65 & \textbf{4.65} & 4.31 & 4.88 & 3.46 & 4.70 & 4.35 \\
Qwen-Image-Edit-2509~\citep{wu2025qwen}     & 20B       & 4.32 & 4.36 & 4.04 & 4.64 & 4.52 & \underline{4.37} & 4.84 & 3.39 & 4.71 & 4.35 \\
Z-Image-Edit~\citep{cai2025z}             & 6B        & 4.40 & 4.14 & \textbf{4.30} & 4.57 & 4.13 & 4.14 & 4.85 & 3.63 & 4.50 & 4.30 \\
Qwen-Image-Edit~\citep{wu2025qwen}          & 20B       & 4.38 & 4.16 & 3.43 & 4.66 & 4.14 & \textbf{4.38} & 4.81 & \underline{3.82} & 4.69 & 4.27 \\
FLUX.1 Kontext [Dev]~\citep{labs2025flux1kontextflowmatching}     & 12B       & 4.12 & 3.80 & 2.04 & 4.22 & 3.09 & 3.97 & 4.51 & 3.35 & 4.25 & 3.71 \\
OmniGen2~\citep{wu2025omnigen2}                 & 4B        & 3.57 & 3.06 & 1.77 & 3.74 & 3.20 & 3.57 & 4.81 & 2.52 & 4.68 & 3.44 \\
Step1X-Edit~\citep{liu2025step1x}               & 12B       & 3.88 & 3.14 & 1.76 & 3.40 & 2.41 & 3.16 & 4.63 & 2.64 & 2.52 & 3.06 \\
OmniGen~\citep{xiao2025omnigen}                   & 3.8B      & 3.47 & 3.04 & 1.71 & 2.94 & 2.43 & 3.21 & 4.19 & 2.24 & 3.38 & 2.96 \\
\midrule
\multicolumn{12}{c}{\emph{Open-source Unified Multimodal Models}} \\
\midrule
Emu3.5~\citep{cui2025emu3}                   & 32B       & \textbf{4.61} & \underline{4.32} & \textbf{3.96} & \textbf{4.84} & \textbf{4.58} & \textbf{4.35} & 4.79 & \textbf{3.69} & \underline{4.57} & \textbf{4.41} \\
Ovis-U1~\citep{wang2025ovis}                  & 1.2B      & 3.99 & 3.73 & 2.66 & 4.38 & 4.15 & 4.05 & \textbf{4.86} & \underline{3.43} & \textbf{4.68} & 3.97 \\
SenseNova-U1~\citep{sensenova2026u1}             & 8B        & 3.83 & 4.15 & 3.12 & 4.32 & 3.26 & 4.18 & \underline{4.85} & 3.03 & 4.41 & 3.90 \\
InternVL-U (w/ CoT)~\citep{tian2026internvl}      & 1.7B      & 4.24 & 3.80 & 2.58 & 4.36 & 3.51 & 3.92 & 4.69 & 3.00 & 4.31 & 3.82 \\
InternVL-U~\citep{tian2026internvl}               & 1.7B      & 4.13 & 3.40 & 2.27 & 4.13 & 3.39 & 3.84 & 4.77 & 3.03 & 4.05 & 3.67 \\
UniWorld-V1~\citep{lin2025uniworld}              & 12B       & 3.82 & 3.64 & 2.27 & 3.47 & 3.24 & 2.99 & 4.21 & 2.96 & 2.74 & 3.26 \\
BAGEL~\citep{deng2025emerging}                     & 7B        & 3.56 & 3.31 & 1.70 & 3.30 & 2.62 & 3.24 & 4.49 & 2.38 & 4.17 & 3.20 \\
\textbf{UniSpace (Ours)}    & 8B        & \underline{4.53} & \textbf{4.38} & \underline{3.61} & \underline{4.67} & \underline{4.42} & \underline{4.23} & 4.55 & 2.70 & 4.47 & \underline{4.28} \\
\bottomrule
\end{tabular}%
}
\end{table}

\begin{table}[t]
\centering
\caption{Additional image-editing evaluation on GEdit. Reference
GEdit-Bench-EN results are reproduced from SenseNova-U1~\citep{sensenova2026u1}; BAGEL-CN is reproduced
from the BAGEL~\citep{deng2025emerging} report; UniSpace is evaluated with GPT-4o. Avg. is computed over
English and Chinese overall scores when both are available. All metrics are
higher-is-better.}
\label{tab:gedit}
\setlength{\tabcolsep}{3.4pt}
\renewcommand{\arraystretch}{1.12}
\resizebox{\textwidth}{!}{%
\begin{tabular}{lcccccccc}
\toprule
& & \multicolumn{3}{c}{GEdit-Bench-EN} & \multicolumn{3}{c}{GEdit-Bench-CN} & \\
\cmidrule(lr){3-5}\cmidrule(lr){6-8}
Model & \# Params & SC & PQ & Overall$\uparrow$ & SC & PQ & Overall$\uparrow$ & Avg.$\uparrow$ \\
\midrule
\multicolumn{9}{c}{\emph{Specialized Image Editing Models}} \\
\midrule
Qwen-Image-Edit-2511~\citep{wu2025qwen} & 20B & \textbf{8.30} & \textbf{8.20} & \textbf{7.88} & -- & -- & -- & -- \\
LongCat-Image-Edit~\citep{team2025longcat}   & 6B  & \underline{8.13} & \underline{8.18} & \underline{7.75} & -- & -- & -- & -- \\
Z-Image-Edit~\citep{cai2025z}         & 6B  & 8.11 & 7.72 & 7.57 & -- & -- & -- & -- \\
Qwen-Image-Edit~\citep{wu2025qwen}      & 20B & 8.00 & 7.86 & 7.56 & -- & -- & -- & -- \\
\midrule
\multicolumn{9}{c}{\emph{Unified Multimodal Models}} \\
\midrule
Emu3.5~\citep{cui2025emu3}               & 32B & 8.11 & \textbf{7.70} & \textbf{7.59} & -- & -- & -- & -- \\
SenseNova-U1~\citep{sensenova2026u1}         & 8B  & \underline{8.27} & \underline{7.49} & \underline{7.47} & -- & -- & -- & -- \\
BAGEL~\citep{deng2025emerging}                & 7B  & 7.36 & 6.83 & 6.52 & \underline{7.34} & \underline{6.85} & \underline{6.50} & \underline{6.51} \\
\textbf{UniSpace (Ours)} & 8B & \textbf{8.29} & 7.06 & 7.41 & \textbf{8.27} & \textbf{7.00} & \textbf{7.38} & \textbf{7.395} \\
\bottomrule
\end{tabular}%
}
\end{table}

\subsection{Scaling Unified Multimodal Modeling}
\label{sec:unispace-experiments}

We next study whether the proposed unified representation can be scaled from
controlled representation learning to a full multimodal system. UniSpace is not
intended to isolate the representation alone: it combines the frozen unified
encoder--decoder with an 8B Mixture-of-Transformer-Experts backbone and is
trained on large-scale internal generation, editing, and understanding data.
We therefore use UniSpace as a system-level validation. The central question is
whether a single ViT-based visual space can replace the conventional separation
between semantic image tokens and VAE latents while still supporting practical
generation and editing.

\subsubsection{Image Editing}

We evaluate UniSpace on two image-editing benchmarks, ImgEdit~\citep{ye2025imgedit}
and GEdit~\citep{sensenova2026u1}. ImgEdit serves as our primary benchmark for
system-level comparison, while GEdit provides a complementary evaluation of
instruction following and perceptual quality under both English and Chinese
prompts. 

\paragraph{ImgEdit.}
ImgEdit covers nine editing categories, ranging from local content manipulation
to style transfer and hybrid editing. As shown in Table~\ref{tab:imgedit},
UniSpace achieves an overall score of $4.28$ at the 8B scale. It substantially
outperforms comparable-scale unified models, including SenseNova-U1 ($3.90$)
and BAGEL ($3.20$), while approaching the $4.41$ score of the substantially
larger 32B Emu3.5. These results demonstrate the strong editing capability of UniSpace at the 8B
scale. Despite using substantially fewer parameters than Emu3.5, UniSpace
achieves a comparable overall score, while clearly outperforming other
unified models of similar scale. This suggests that the unified representation
provides sufficient semantic and reconstruction information for precise
instruction-based image editing.

\paragraph{GEdit.}
We further evaluate UniSpace on GEdit. As GEdit serves as a complementary
benchmark, we report aggregate results and compare UniSpace with other
representative methods in Table~\ref{tab:gedit}. UniSpace achieves overall
scores of $7.41$ and $7.38$ on the English and Chinese full sets,
respectively, yielding a bilingual average of $7.39$. It outperforms BAGEL and
remains close to the dense SenseNova-U1 8B model. The detailed results show
that UniSpace achieves strong semantic consistency, while its perceptual
quality remains relatively weaker, particularly on compositional editing and
other perceptually sensitive cases.

\paragraph{Qualitative Results.}

\begin{figure}[p]
\centering
\includegraphics[width=\linewidth,height=0.88\textheight,keepaspectratio]{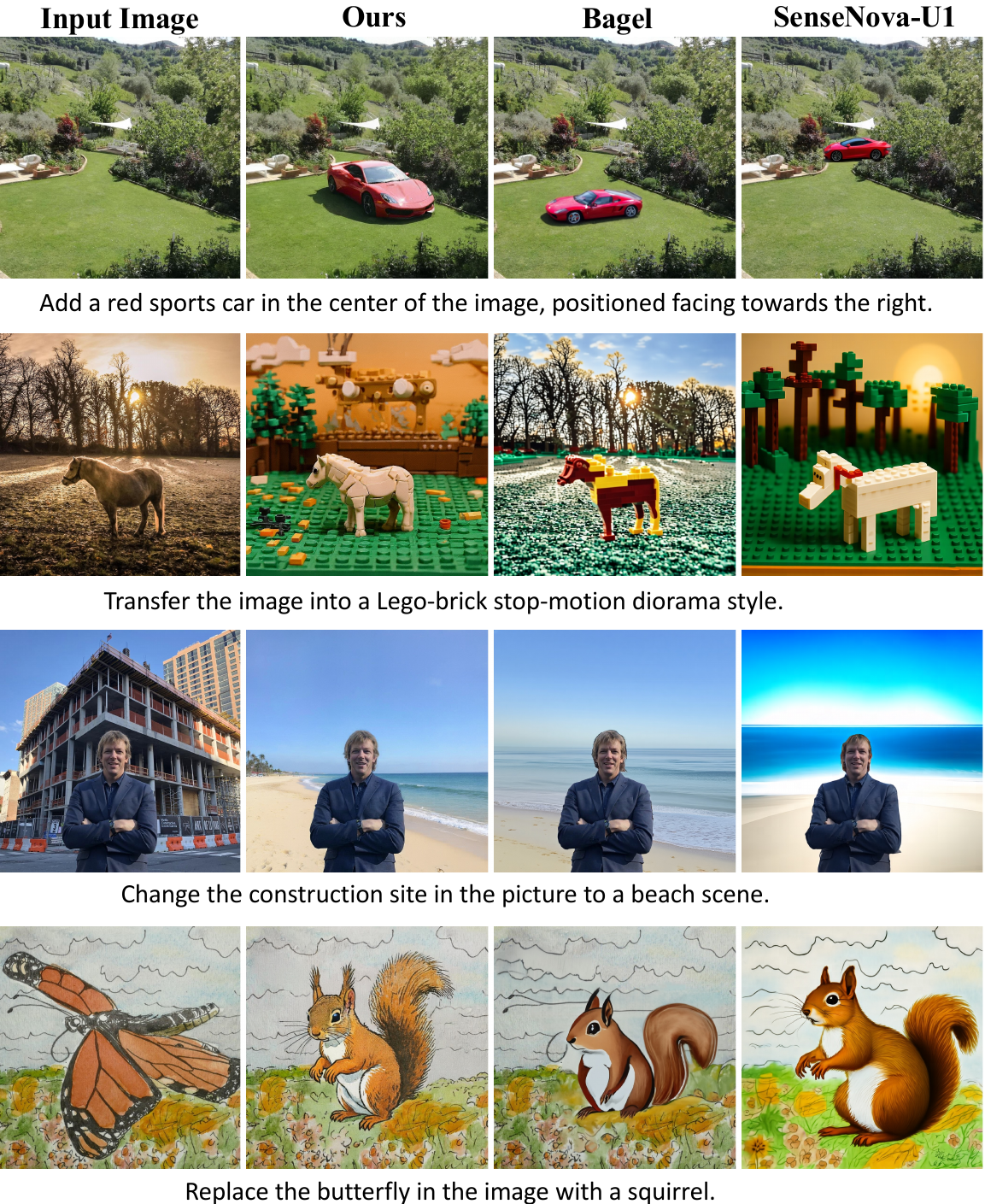}
\caption{Qualitative image editing comparison.
  Columns (left to right): input image, Ours (UniSpace), BAGEL, SenseNova-U1.
  Rows (top to bottom): object addition, style transfer, background replacement, object replacement.}
\label{fig:imageedit-qual}
\end{figure}

\begin{figure}[p]
\centering
\includegraphics[width=\linewidth,height=0.88\textheight,keepaspectratio]{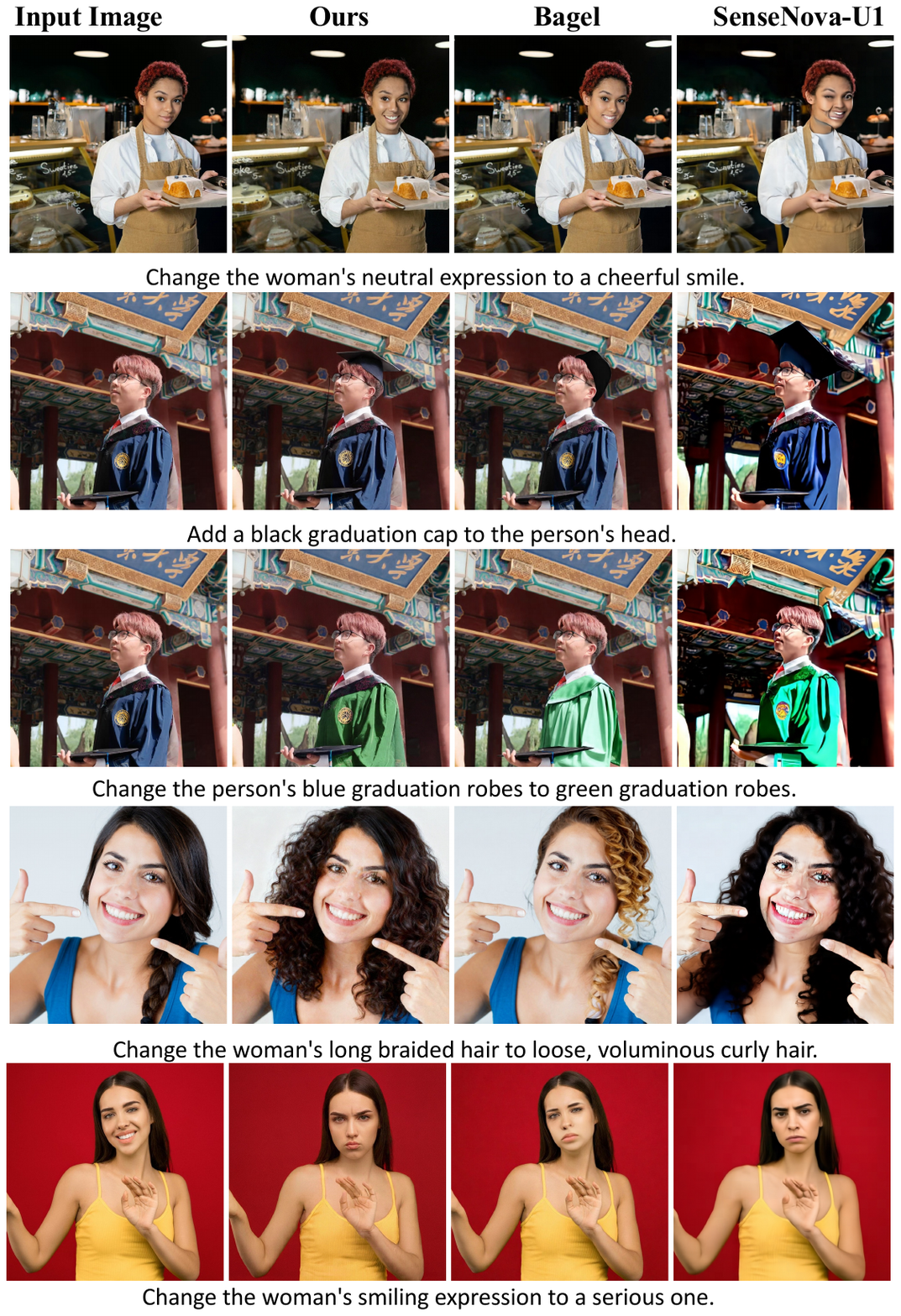}
\caption{Qualitative comparison on human-centric image editing.
  Columns (left to right): input image, Ours (UniSpace), BAGEL, SenseNova-U1.
  Rows: expression change (neutral$\to$smile), accessory addition (graduation cap),
  clothing recoloring (blue$\to$green robes), hairstyle transfer (braids$\to$curly),
  expression change (smile$\to$serious).}
\label{fig:humanedit-qual}
\end{figure}

Figure~\ref{fig:imageedit-qual} presents qualitative comparisons on four
representative editing tasks: object addition, style transfer, scene
replacement, and object substitution. UniSpace consistently follows the
editing instructions while preserving the identity and structure of unrelated
image regions. In contrast, BAGEL and SenseNova-U1 tend to produce artifacts or
exhibit incomplete instruction adherence in challenging cases, particularly for
fine-grained style transfer and multi-object compositional editing. Figure~\ref{fig:humanedit-qual} shows additional results on human-centric
editing tasks, including expression change, accessory addition, clothing color
modification, and hairstyle transfer. These examples require fine-grained
localized editing while preserving identity and unrelated appearance
attributes. UniSpace performs these edits while maintaining the overall
appearance and structural consistency of the input subjects, further
supporting its ability to combine instruction following with localized visual
editing.

\subsubsection{Text-to-Image Generation}

We evaluate UniSpace on three text-to-image generation benchmarks:
GenEval~\citep{ghosh2023geneval}, OneIG-Bench~\citep{chang2025oneigbench},
and DPG-Bench~\citep{hu2024ella}. GenEval evaluates object-centric
compositional alignment, OneIG-Bench measures fine-grained generation across
English and Chinese prompts, and DPG-Bench evaluates dense-prompt following
across global, entity, attribute, and relational constraints. We adopt the
comparison sets collected by SenseNova-U1~\citep{sensenova2026u1} and report the
results in Tables~\ref{tab:unispace-geneval}--\ref{tab:unispace-dpg}.

\paragraph{GenEval.}
As shown in Table~\ref{tab:unispace-geneval}, UniSpace achieves an overall
score of $0.84$.  UniSpace performs well on single-object generation, two-object
generation, positional alignment, and attribute binding, although its counting
and color-binding scores remain relatively weaker. 

\paragraph{OneIG-Bench.}
On OneIG-Bench, UniSpace obtains overall scores of $0.561$ and $0.533$ on the
English and Chinese subsets, respectively, yielding a bilingual average of
$0.547$. As shown in Tables~\ref{tab:unispace-oneig-en} and
\ref{tab:unispace-oneig-zh}, UniSpace achieves the strongest bilingual average
among the compared unified multimodal models, slightly exceeding Emu3.5
($0.546$) and SenseNova-U1 ($0.542$). It also obtains the highest style scores
among the compared methods on both the English and Chinese subsets, with scores
of $0.467$ and $0.455$, respectively. These results demonstrate that UniSpace
remains competitive in multilingual text-to-image generation, particularly in
style control and fine-grained visual synthesis.

\paragraph{DPG-Bench.}
On DPG-Bench, UniSpace achieves an overall score of $86.49$. It outperforms
BAGEL ($85.07$), Show-o2 ($86.14$), and most other listed unified multimodal
models. UniSpace achieves the highest relation score among all compared models,
reaching $94.97$, and obtains a strong entity score of $92.26$. These results
indicate that UniSpace can effectively follow dense prompts, particularly those
involving relational and entity-level constraints, although its global and
attribute scores remain below those of the strongest text-to-image systems.

\paragraph{Qualitative Results.}

\begin{figure}[p]
\centering
\includegraphics[height=0.9\textheight,keepaspectratio]{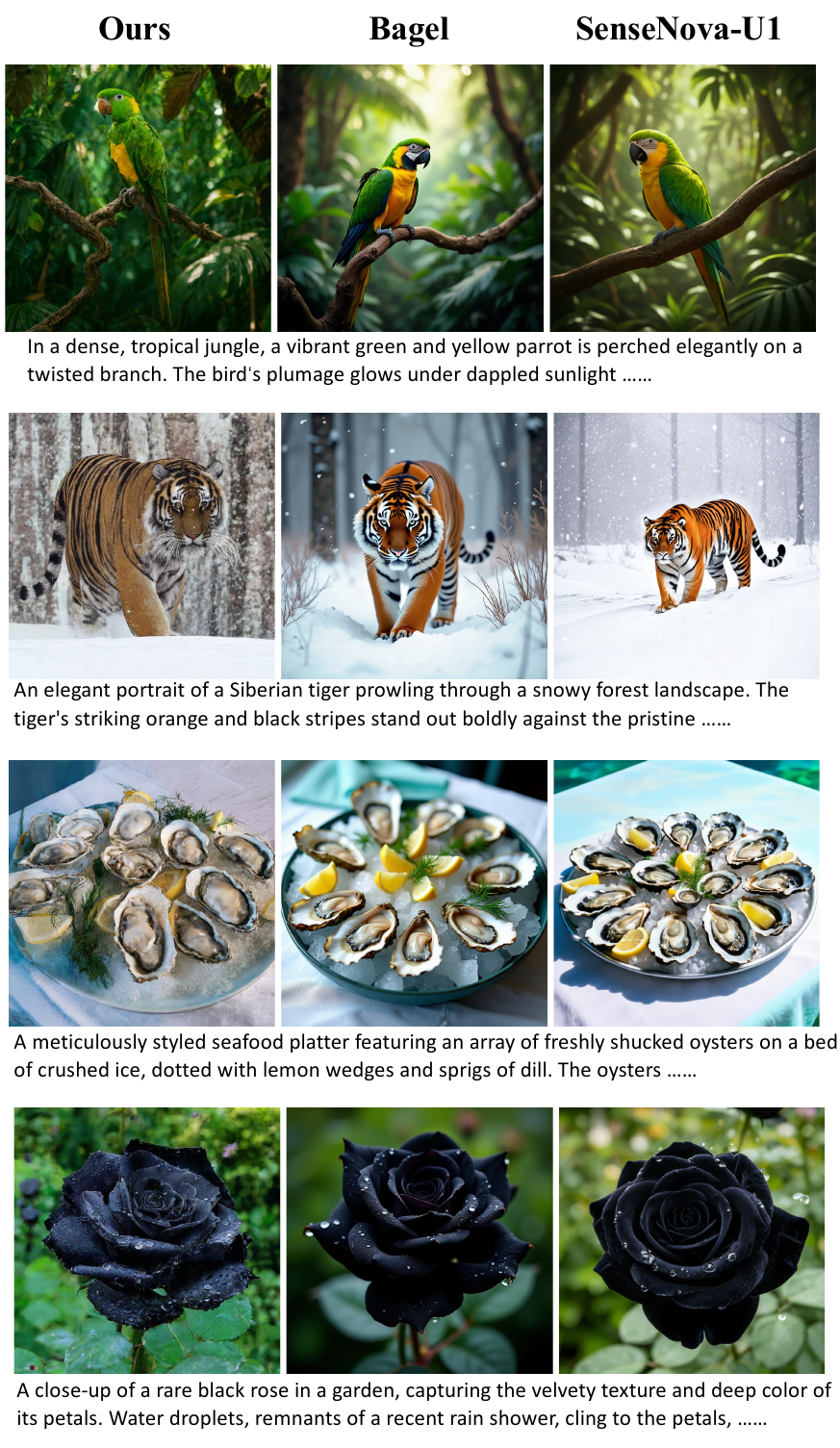}
\caption{Qualitative text-to-image comparison. Columns show Ours (UniSpace),
BAGEL, and SenseNova-U1. Each row is generated from the prompt shown below the
corresponding images.}
\label{fig:t2i-qual}
\end{figure}

\begin{figure}[p]
\centering
\includegraphics[height=0.9\textheight,keepaspectratio]{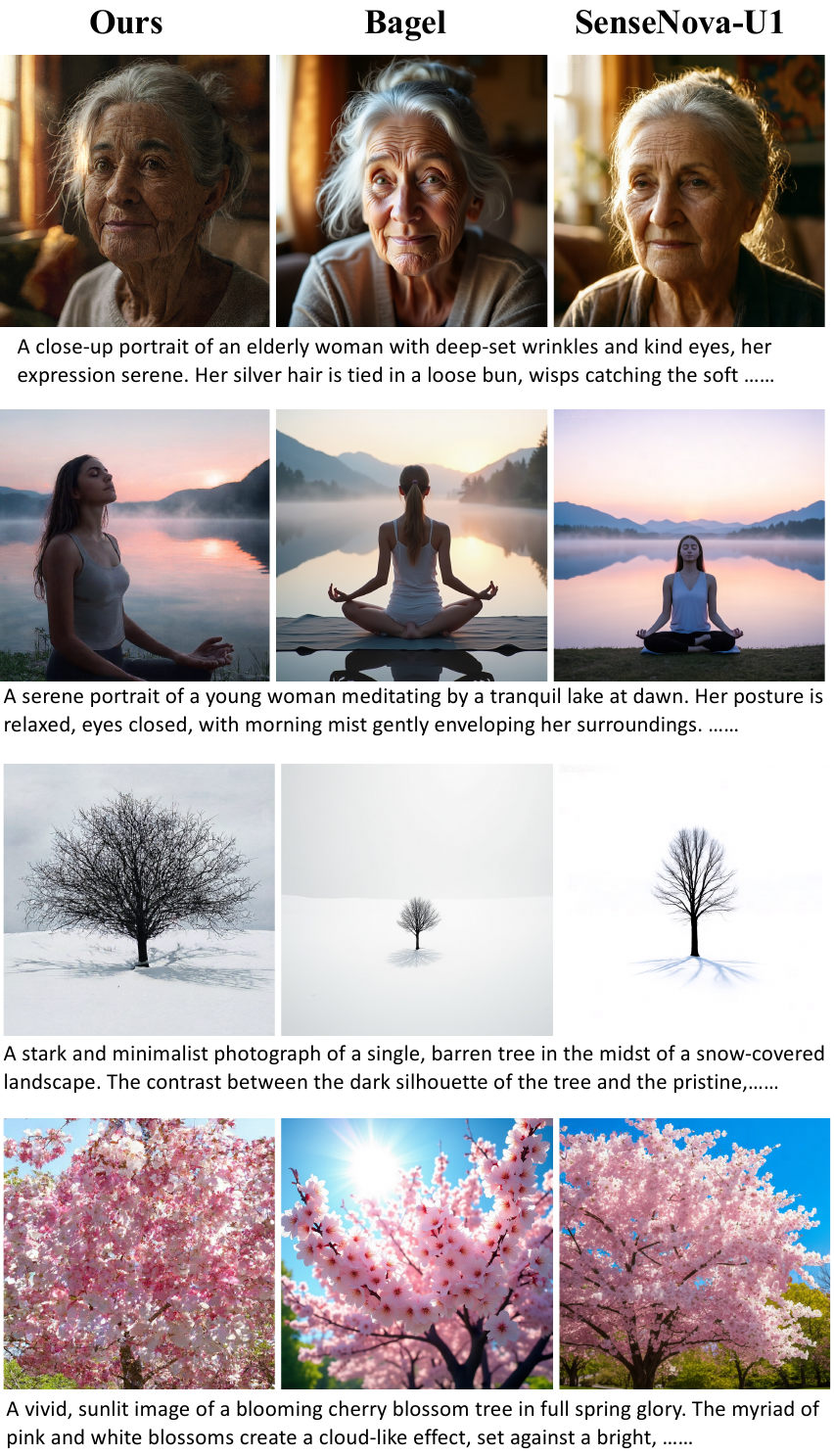}
\caption{Additional qualitative text-to-image comparison. Columns show Ours
(UniSpace), BAGEL, and SenseNova-U1. Each row is generated from the prompt shown
below the corresponding images.}
\label{fig:t2i-qual-additional}
\end{figure}

Figures~\ref{fig:t2i-qual} and~\ref{fig:t2i-qual-additional} present qualitative
text-to-image comparisons between UniSpace, BAGEL, and SenseNova-U1. UniSpace
generally produces images that are well aligned with the input prompts while
maintaining coherent object layouts, attributes, and visual styles. Compared
with the other unified models, its generations exhibit a more realistic
appearance and richer visual details, especially in complex scenes and
fine-grained visual content. These results provide qualitative evidence that
the unified representation supports both prompt alignment and high-fidelity
visual synthesis.

\begin{table}[t]
\centering
\caption{Text-to-image generation on GenEval. SO, TO, CT, CL, POS, and ATTR
denote single object, two objects, counting, colors, position, and attribute
binding, respectively. All metrics are higher-is-better.}
\label{tab:unispace-geneval}
\setlength{\tabcolsep}{3.4pt}
\renewcommand{\arraystretch}{1.10}
\resizebox{\textwidth}{!}{%
\begin{tabular}{lcrrrrrrr}
\toprule
Model & \# Params & SO & TO & CT & CL & POS & ATTR & Overall$\uparrow$ \\
\midrule
\multicolumn{9}{c}{\emph{Open-source Generation Models}} \\
\midrule
Qwen-Image~\citep{wu2025qwen}           & 20B & \underline{0.99} & \underline{0.92} & \textbf{0.89} & 0.88 & \textbf{0.76} & \textbf{0.77} & \textbf{0.87} \\
Z-Image~\citep{cai2025z}              & 6B  & \textbf{1.00} & \textbf{0.94} & \underline{0.78} & \textbf{0.93} & \underline{0.62} & \textbf{0.77} & \underline{0.84} \\
SD3-Medium~\citep{rombach2021highresolution}           & 2B  & \underline{0.99} & \textbf{0.94} & 0.72 & \underline{0.89} & 0.33 & \underline{0.60} & 0.74 \\
FLUX.1-dev~\citep{flux2024}           & 12B & 0.98 & 0.81 & 0.74 & 0.79 & 0.22 & 0.45 & 0.66 \\
\midrule
\multicolumn{9}{c}{\emph{Open-source Unified Multimodal Models}} \\
\midrule
SenseNova-U1~\citep{sensenova2026u1}         & 8B  & \textbf{1.00} & \underline{0.96} & \textbf{0.92} & \textbf{0.92} & \textbf{0.91} & 0.76 & \textbf{0.91} \\
InternVL-U~\citep{tian2026internvl}           & 1.7B& \underline{0.99} & 0.94 & 0.74 & \underline{0.91} & 0.77 & 0.74 & 0.85 \\
BAGEL~\citep{deng2025emerging}                & 7B  & \underline{0.99} & 0.94 & 0.81 & 0.88 & 0.64 & 0.63 & 0.82 \\
Janus-Pro~\citep{chen2025janus}            & 7B  & \underline{0.99} & 0.89 & 0.59 & 0.90 & 0.79 & 0.66 & 0.80 \\
OmniGen2~\citep{wu2025omnigen2}             & 4B  & \textbf{1.00} & 0.95 & 0.64 & 0.88 & 0.55 & 0.76 & 0.80 \\
UniWorld-V1~\citep{lin2025uniworld}          & 12B & \underline{0.99} & 0.93 & 0.79 & 0.89 & 0.49 & 0.70 & 0.80 \\
Show-o2~\citep{xie2026show}              & 7B  & \textbf{1.00} & 0.87 & 0.58 & \textbf{0.92} & 0.52 & 0.62 & 0.76 \\
Emu3.5~\citep{cui2025emu3}               & 32B & --   & --   & --   & --   & --   & --   & 0.73 \\
\textbf{UniSpace (Ours)} & 8B & 0.98 & 0.92 & 0.69 & 0.88 & 0.83 & 0.73 & 0.84 \\
\bottomrule
\end{tabular}%
}
\end{table}

\begin{table}[t]
\centering
\caption{Text-to-image generation on OneIG-Bench (English).}
\label{tab:unispace-oneig-en}
\setlength{\tabcolsep}{5.0pt}
\renewcommand{\arraystretch}{1.10}
\resizebox{\textwidth}{!}{%
\begin{tabular}{lcrrrrrrr}
\toprule
Model & \# Params & Align. & Text & Reason. & Style & Div. & Overall$\uparrow$ \\
\midrule
\multicolumn{8}{c}{\emph{Closed-source Models}} \\
\midrule
Gemini-2.5-Flash-Image & --  & \textbf{0.878} & \textbf{0.894} & \textbf{0.346} & \underline{0.450} & \underline{0.182} & \textbf{0.550} \\
GPT-Image-1            & --  & \underline{0.851} & 0.857 & \underline{0.345} & \textbf{0.462} & 0.151 & \underline{0.533} \\
Seedream 3.0           & --  & 0.818 & \underline{0.865} & 0.275 & 0.413 & \textbf{0.277} & 0.530 \\
\midrule
\multicolumn{8}{c}{\emph{Open-source Generation Models}} \\
\midrule
Qwen-Image~\citep{wu2025qwen}             & 20B & 0.882 & 0.891 & 0.306 & 0.418 & 0.197 & 0.539 \\
\midrule
\multicolumn{8}{c}{\emph{Open-source Unified Multimodal Models}} \\
\midrule
Emu3.5~\citep{cui2025emu3}                 & 32B & \textbf{0.902} & \textbf{0.994} & \textbf{0.345} & 0.427 & 0.151 & \textbf{0.564} \\
SenseNova-U1~\citep{sensenova2026u1}           & 8B  & \underline{0.882} & \underline{0.969} & \underline{0.330} & 0.396 & 0.166 & 0.549 \\
BAGEL~\citep{deng2025emerging}                  & 7B  & 0.769 & 0.244 & 0.173 & 0.367 & \underline{0.251} & 0.361 \\
Janus-Pro~\citep{chen2025janus}              & 7B  & 0.553 & 0.001 & 0.139 & 0.276 & \textbf{0.365} & 0.267 \\
\textbf{UniSpace (Ours)}  & 8B  & 0.860 & 0.937 & 0.311 & \textbf{0.467} & 0.233 & \underline{0.561} \\
\bottomrule
\end{tabular}%
}
\end{table}

\begin{table}[t]
\centering
\caption{Text-to-image generation on OneIG-Bench (Chinese). Baseline results
are reproduced from SenseNova-U1~\citep{sensenova2026u1}. All metrics are
higher-is-better.}
\label{tab:unispace-oneig-zh}
\setlength{\tabcolsep}{5.0pt}
\renewcommand{\arraystretch}{1.10}
\resizebox{\textwidth}{!}{%
\begin{tabular}{lcrrrrrrr}
\toprule
Model & \# Params & Align. & Text & Reason. & Style & Div. & Overall$\uparrow$ \\
\midrule
\multicolumn{8}{c}{\emph{Closed-source Models}} \\
\midrule
Gemini-2.5-Flash-Image & --  & \textbf{0.825} & 0.276 & \underline{0.298} & \underline{0.427} & \underline{0.198} & 0.337 \\
GPT-Image-1            & --  & \underline{0.812} & 0.650 & \textbf{0.300} & \textbf{0.449} & 0.159 & \underline{0.474} \\
Seedream 3.0           & --  & 0.793 & \textbf{0.928} & 0.281 & 0.397 & \textbf{0.243} & \textbf{0.528} \\
\midrule
\multicolumn{8}{c}{\emph{Open-source Generation Models}} \\
\midrule
Qwen-Image~\citep{wu2025qwen}             & 20B & 0.825 & 0.963 & 0.267 & 0.405 & 0.279 & 0.548 \\
\midrule
\multicolumn{8}{c}{\emph{Open-source Unified Multimodal Models}} \\
\midrule
Emu3.5~\citep{cui2025emu3}                 & 32B & \textbf{0.853} & \underline{0.941} & \underline{0.300} & 0.386 & 0.166 & 0.529 \\
SenseNova-U1~\citep{sensenova2026u1}           & 8B  & \underline{0.826} & \textbf{0.977} & \textbf{0.303} & \underline{0.392} & 0.176 & \textbf{0.535} \\
BAGEL~\citep{deng2025emerging}                  & 7B  & 0.672 & 0.365 & 0.186 & 0.357 & \underline{0.268} & 0.370 \\
Janus-Pro~\citep{chen2025janus}              & 7B  & 0.324 & 0.148 & 0.104 & 0.264 & \textbf{0.358} & 0.240 \\
\textbf{UniSpace (Ours)}  & 8B  & 0.807 & 0.881 & 0.276 & \textbf{0.455} & 0.244 & \underline{0.533} \\
\bottomrule
\end{tabular}%
}
\end{table}

\begin{table}[t]
\centering
\caption{Dense-prompt following performance on DPG-Bench. All metrics are
higher-is-better.}
\label{tab:unispace-dpg}
\setlength{\tabcolsep}{4.0pt}
\renewcommand{\arraystretch}{1.10}
\resizebox{\textwidth}{!}{%
\begin{tabular}{lccccccc}
\toprule
Model & \# Params & Global & Entity & Attribute & Relation & Other & Overall$\uparrow$ \\
\midrule
\multicolumn{8}{c}{\emph{Closed-source Models}} \\
\midrule
Seedream 4.5          & --  & \underline{89.24} & \textbf{94.30} & \textbf{92.14} & 92.23 & \textbf{93.83} & \textbf{88.63} \\
Nano-Banana-Pro       & --  & \textbf{91.00} & \underline{92.85} & \underline{91.56} & \underline{92.39} & 89.93 & \underline{87.16} \\
GPT-Image-1           & --  & 88.89 & 88.94 & 89.84 & \textbf{92.63} & \underline{90.96} & 85.15 \\
\midrule
\multicolumn{8}{c}{\emph{Open-source Generation Models}} \\
\midrule
Qwen-Image~\citep{wu2025qwen}            & 20B & \underline{91.32} & \textbf{91.56} & \underline{92.02} & \textbf{94.31} & \textbf{92.73} & \textbf{88.32} \\
Z-Image~\citep{cai2025z}               & 6B  & \textbf{93.39} & \underline{91.22} & \textbf{93.16} & \underline{92.22} & \underline{91.52} & \underline{88.14} \\
SD3-Medium~\citep{rombach2021highresolution}            & 2B  & 87.90 & 91.01 & 88.83 & 80.70 & 88.68 & 84.08 \\
FLUX.1-dev~\citep{flux2024}            & 12B & 74.35 & 90.00 & 88.96 & 90.87 & 88.33 & 83.84 \\
\midrule
\multicolumn{8}{c}{\emph{Open-source Unified Multimodal Models}} \\
\midrule
SenseNova-U1~\citep{sensenova2026u1}          & 8B  & 88.74 & 90.90 & \textbf{92.43} & 92.43 & \textbf{92.50} & \textbf{87.78} \\
Tuna~\citep{liu2026tuna}                  & 7B  & \underline{90.42} & \underline{91.68} & 90.94 & 91.87 & \underline{90.73} & \underline{86.76} \\
NEO-unify~\citep{sensenova2026u1}             & 8B  & \textbf{91.00} & 91.53 & \underline{92.06} & \underline{94.14} & 90.43 & 86.71 \\
Show-o2~\citep{xie2026show}               & 7B  & --    & --    & --    & --    & --    & 86.14 \\
InternVL-U~\citep{tian2026internvl}            & 1.7B& 90.39 & 90.78 & 90.68 & 90.29 & 88.77 & 85.18 \\
BAGEL~\citep{deng2025emerging}                 & 7B  & 88.94 & 90.37 & 91.29 & 90.82 & 88.67 & 85.07 \\
Janus-Pro~\citep{chen2025janus}             & 7B  & 86.90 & 88.90 & 89.40 & 89.32 & 89.48 & 84.19 \\
Ovis-U1~\citep{wang2025ovis}               & 1.2B& 82.37 & 90.08 & 88.68 & 93.35 & 85.20 & 83.72 \\
OmniGen2~\citep{wu2025omnigen2}              & 4B  & 88.81 & 88.83 & 90.18 & 89.37 & 90.27 & 83.57 \\
UniWorld-V1~\citep{lin2025uniworld}           & 12B & 83.64 & 88.39 & 88.44 & 89.27 & 87.22 & 81.38 \\
\textbf{UniSpace (Ours)} & 8B & 84.80 & \textbf{92.26} & 90.00 & \textbf{94.97} & 88.80 & 86.49 \\
\bottomrule
\end{tabular}%
}
\end{table}


\subsubsection{Unified Capability Evaluation}

The same model also retains an understanding pathway, but we do not use UniSpace's
system-level understanding scores as the main evidence for this paper. UniSpace is
optimized primarily for generation and editing, and its post-training recipe is
not designed to maximize standard VLM benchmarks. Instead, the understanding
capability of the unified representation itself is validated in the controlled
experiments in Table~\ref{tab:multimodal-understanding}, where all tasks use
the complete unified representation $T_u$.
\section{Limitation}
\label{sec:limitation}
Our results establish the viability of a shared visual representation, but do not imply uniform improvements across all downstream tasks. UniSpace is primarily optimized for generation and editing, and its system-level understanding performance still lags behind that of dedicated vision--language models, despite controlled experiments showing that the representation itself largely preserves the capabilities of the original semantic encoder. Moreover, because the encoder serves as a unified representation for both understanding and generation, it cannot be freely updated during training on understanding data without potentially compromising its generative capabilities. This architectural constraint limits further improvements in visual understanding. An important direction for future work is therefore to explore how the unified representation and UniMM can be jointly optimized while maintaining a balance between understanding and generation.

\section{Conclusion}
\label{sec:conclusion}

We presented Patch Reparameterization, a minimal adaptation that turns a
pretrained semantic ViT into a unified visual tokenizer without modifying its
Transformer blocks. By retaining the original semantic pathway, introducing a
reconstruction-aware patch pathway, and explicitly factorizing their outputs,
the proposed representation supports multimodal understanding, high-fidelity
reconstruction, and generation within a single frozen parameter space. More
broadly, our findings suggest that unifying visual capabilities need not begin
with training a new vision backbone: changing how information enters an
existing semantic encoder can be sufficient to expose a substantially broader
range of visual information.

Scaling the same frozen representation to UniSpace further shows that a single
visual space can replace the conventional ViT--VAE dual interface in a large
multimodal system. Within an 8B mixture-of-experts model, UniSpace supports multimodal understanding, high-quality image editing, and practical
text-to-image generation. In particular, it achieves high-quality image
editing and competitive text-to-image generation while maintaining a compact
8B model scale, demonstrating the practicality of a unified visual space for
large-scale multimodal systems. We release the complete UniSpace system and
hope that it will facilitate further research on unified multimodal models.

\bibliography{refs}

@article{deng2025emerging,
  title={Emerging Properties in Unified Multimodal Pretraining},
  author={Deng, Chaorui and Zhu, Deyao and Li, Kunchang and Gou, Chenhui and Li, Feng and Wang, Zeyu and Zhong, Shu and Yu, Weihao and Nie, Xiaonan and Song, Ziang and Shi, Guang and Fan, Haoqi},
  journal={arXiv preprint arXiv:2505.14683},
  year={2025}
}

@article{sensenova2026u1,
  title={SenseNova-U1: Unifying Multimodal Understanding and Generation with NEO-unify Architecture},
  author={{SenseNova-U1 Team}},
  journal={arXiv preprint arXiv:2605.12500},
  year={2026}
}

@article{ye2025imgedit,
  title={ImgEdit: A Unified Image Editing Dataset and Benchmark},
  author={Ye, Yang and He, Xianyi and Li, Zongjian and Lin, Bin and Yuan, Shenghai and Yan, Zhiyuan and Hou, Bohan and Yuan, Li},
  journal={arXiv preprint arXiv:2505.20275},
  year={2025}
}

@inproceedings{ghosh2023geneval,
  title={GenEval: An Object-Focused Framework for Evaluating Text-to-Image Alignment},
  author={Ghosh, Dhruba and Hajishirzi, Hannaneh and Schmidt, Ludwig},
  booktitle={Advances in Neural Information Processing Systems},
  year={2023}
}

@article{hu2024ella,
  title={ELLA: Equip Diffusion Models with LLM for Enhanced Semantic Alignment},
  author={Hu, Xiwei and Wang, Rui and Fang, Yixiao and Fu, Bin and Cheng, Pei and Yu, Gang},
  journal={arXiv preprint arXiv:2403.05135},
  year={2024}
}

@article{chang2025oneigbench,
  title={OneIG-Bench: Omni-Dimensional Nuanced Evaluation for Image Generation},
  author={Chang, Jingjing and Fang, Yixiao and Xing, Peng and Wu, Shuhan and Cheng, Wei and Wang, Rui and Zeng, Xianfang and Yu, Gang and Chen, Hai-Bao},
  journal={arXiv preprint arXiv:2506.07977},
  year={2025}
}

@article{tschannen2025siglip2,
  title={SigLIP 2: Multilingual Vision-Language Encoders with Improved Semantic Understanding, Localization, and Dense Features},
  author={Tschannen, Michael and Gritsenko, Alexey and Wang, Xiao and Naeem, Muhammad Ferjad and Alabdulmohsin, Ibrahim and Parthasarathy, Nikhil and Evans, Talfan and Beyer, Lucas and Xia, Ye and Mustafa, Basil and others},
  journal={arXiv preprint arXiv:2502.14786},
  year={2025}
}

@inproceedings{radford2021learning,
  title={Learning Transferable Visual Models From Natural Language Supervision},
  author={Radford, Alec and Kim, Jong Wook and Hallacy, Chris and Ramesh, Aditya and Goh, Gabriel and Agarwal, Sandhini and Sastry, Girish and Askell, Amanda and Mishkin, Pamela and Clark, Jack and Krueger, Gretchen and Sutskever, Ilya},
  booktitle={International Conference on Machine Learning},
  pages={8748--8763},
  year={2021},
  organization={PMLR}
}

@inproceedings{oquab2023dinov2,
  title={DINOv2: Learning Robust Visual Features without Supervision},
  author={Oquab, Maxime and Darcet, Timoth{\'e}e and Moutakanni, Th{\'e}o and Vo, Huy and Szafraniec, Marc and Khalidov, Vasil and Fernandez, Pierre and Haziza, Daniel and Massa, Francisco and El-Nouby, Alaaeldin and others},
  booktitle={Transactions on Machine Learning Research},
  year={2024}
}

@article{zheng2025rae,
  title={Diffusion Transformers with Representation Autoencoders},
  author={Zheng, Boyang and Ma, Nanye and Tong, Shengbang and Xie, Saining},
  journal={arXiv preprint arXiv:2510.11690},
  year={2025}
}

@inproceedings{yao2025reconstruction,
  title={Reconstruction vs. generation: Taming optimization dilemma in latent diffusion models},
  author={Yao, Jingfeng and Yang, Bin and Wang, Xinggang},
  booktitle={Proceedings of the Computer Vision and Pattern Recognition Conference},
  pages={15703--15712},
  year={2025}
}

@article{yue2025uniflow,
  title={Uniflow: A unified pixel flow tokenizer for visual understanding and generation},
  author={Yue, Zhengrong and Zhang, Haiyu and Zeng, Xiangyu and Chen, Boyu and Wang, Chenting and Zhuang, Shaobin and Dong, Lu and Wang, Yi and Wang, Limin and Wang, Yali},
  journal={arXiv preprint arXiv:2510.10575},
  year={2025}
}

@article{yao2025towards,
  title={Towards Scalable Pre-training of Visual Tokenizers for Generation},
  author={Yao, Jingfeng and Song, Yuda and Zhou, Yucong and Wang, Xinggang},
  journal={arXiv preprint arXiv:2512.13687},
  year={2025}
}

@article{wang2024emu3,
  title={Emu3: Next-token prediction is all you need},
  author={Wang, Xinlong and Zhang, Xiaosong and Luo, Zhengxiong and Sun, Quan and Cui, Yufeng and Wang, Jinsheng and Zhang, Fan and Wang, Yueze and Li, Zhen and Yu, Qiying and others},
  journal={arXiv preprint arXiv:2409.18869},
  year={2024}
}

@article{cui2025emu3,
  title={Emu3. 5: Native multimodal models are world learners},
  author={Cui, Yufeng and Chen, Honghao and Deng, Haoge and Huang, Xu and Li, Xinghang and Liu, Jirong and Liu, Yang and Luo, Zhuoyan and Wang, Jinsheng and Wang, Wenxuan and others},
  journal={arXiv preprint arXiv:2510.26583},
  year={2025}
}

@article{yu2024representation,
  title={Representation alignment for generation: Training diffusion transformers is easier than you think},
  author={Yu, Sihyun and Kwak, Sangkyung and Jang, Huiwon and Jeong, Jongheon and Huang, Jonathan and Shin, Jinwoo and Xie, Saining},
  journal={arXiv preprint arXiv:2410.06940},
  year={2024}
}

@article{singh2026improved,
  title={Improved baselines with representation autoencoders},
  author={Singh, Jaskirat and Zheng, Boyang and Wu, Zongze and Zhang, Richard and Shechtman, Eli and Xie, Saining},
  journal={arXiv preprint arXiv:2605.18324},
  year={2026}
}

@inproceedings{liu2024improved,
  title={Improved baselines with visual instruction tuning},
  author={Liu, Haotian and Li, Chunyuan and Li, Yuheng and Lee, Yong Jae},
  booktitle={Proceedings of the IEEE/CVF conference on computer vision and pattern recognition},
  pages={26296--26306},
  year={2024}
}

@misc{liu2024llavanext,
    title={LLaVA-NeXT: Improved reasoning, OCR, and world knowledge},
    url={https://llava-vl.github.io/blog/2024-01-30-llava-next/},
    author={Liu, Haotian and Li, Chunyuan and Li, Yuheng and Li, Bo and Zhang, Yuanhan and Shen, Sheng and Lee, Yong Jae},
    month={January},
    year={2024}
}

@article{yang2025qwen3,
  title={Qwen3 technical report},
  author={Yang, An and Li, Anfeng and Yang, Baosong and Zhang, Beichen and Hui, Binyuan and Zheng, Bo and Yu, Bowen and Gao, Chang and Huang, Chengen and Lv, Chenxu and others},
  journal={arXiv preprint arXiv:2505.09388},
  year={2025}
}

@article{bai2025qwen3,
  title={Qwen3-vl technical report},
  author={Bai, Shuai and Cai, Yuxuan and Chen, Ruizhe and Chen, Keqin and Chen, Xionghui and Cheng, Zesen and Deng, Lianghao and Ding, Wei and Gao, Chang and Ge, Chunjiang and others},
  journal={arXiv preprint arXiv:2511.21631},
  year={2025}
}

@article{liu2025step1x,
  title={Step1x-edit: A practical framework for general image editing},
  author={Liu, Shiyu and Han, Yucheng and Xing, Peng and Yin, Fukun and Wang, Rui and Cheng, Wei and Liao, Jiaqi and Wang, Yingming and Fu, Honghao and Han, Chunrui and others},
  journal={arXiv preprint arXiv:2504.17761},
  year={2025}
}

@misc{rombach2021highresolution,
      title={High-Resolution Image Synthesis with Latent Diffusion Models},
      author={Robin Rombach and Andreas Blattmann and Dominik Lorenz and Patrick Esser and Björn Ommer},
      year={2021},
      eprint={2112.10752},
      archivePrefix={arXiv},
      primaryClass={cs.CV}
}

@misc{labs2025flux1kontextflowmatching,
      title={FLUX.1 Kontext: Flow Matching for In-Context Image Generation and Editing in Latent Space},
      author={Black Forest Labs and Stephen Batifol and Andreas Blattmann and Frederic Boesel and Saksham Consul and Cyril Diagne and Tim Dockhorn and Jack English and Zion English and Patrick Esser and Sumith Kulal and Kyle Lacey and Yam Levi and Cheng Li and Dominik Lorenz and Jonas Müller and Dustin Podell and Robin Rombach and Harry Saini and Axel Sauer and Luke Smith},
      year={2025},
      eprint={2506.15742},
      archivePrefix={arXiv},
      primaryClass={cs.GR},
      url={https://arxiv.org/abs/2506.15742},
}

@misc{flux2024,
    author={Black Forest Labs},
    title={FLUX},
    year={2024},
    howpublished={\url{https://github.com/black-forest-labs/flux}},
}

@article{wu2025qwen,
  title={Qwen-image technical report},
  author={Wu, Chenfei and Li, Jiahao and Zhou, Jingren and Lin, Junyang and Gao, Kaiyuan and Yan, Kun and Yin, Sheng-ming and Bai, Shuai and Xu, Xiao and Chen, Yilei and others},
  journal={arXiv preprint arXiv:2508.02324},
  year={2025}
}

@article{luo2024open,
  title={Open-magvit2: An open-source project toward democratizing auto-regressive visual generation},
  author={Luo, Zhuoyan and Shi, Fengyuan and Ge, Yixiao and Yang, Yujiu and Wang, Limin and Shan, Ying},
  journal={arXiv preprint arXiv:2409.04410},
  year={2024}
}

@inproceedings{qu2025tokenflow,
  title={Tokenflow: Unified image tokenizer for multimodal understanding and generation},
  author={Qu, Liao and Zhang, Huichao and Liu, Yiheng and Wang, Xu and Jiang, Yi and Gao, Yiming and Ye, Hu and Du, Daniel K and Yuan, Zehuan and Wu, Xinglong},
  booktitle={Proceedings of the Computer Vision and Pattern Recognition Conference},
  pages={2545--2555},
  year={2025}
}

@article{ma2026unitok,
  title={Unitok: A unified tokenizer for visual generation and understanding},
  author={Ma, Chuofan and Jiang, Yi and Wu, Junfeng and Yang, Jihan and Yu, Xin and Yuan, Zehuan and Peng, Bingyue and Qi, Xiaojuan},
  journal={Advances in Neural Information Processing Systems},
  volume={38},
  pages={129274--129297},
  year={2026}
}

@inproceedings{wu2025vila,
  title={Vila-u: a unified foundation model integrating visual understanding and generation},
  author={Wu, Yecheng and Zhang, Zhuoyang and Chen, Junyu and Tang, Haotian and Li, Dacheng and Fang, Yunhao and Zhu, Ligeng and Xie, Enze and Yin, Hongxu and Yi, Li and others},
  booktitle={International Conference on Learning Representations},
  volume={2025},
  pages={93620--93638},
  year={2025}
}

@article{zhao2025qlip,
  title={Qlip: Text-aligned visual tokenization unifies auto-regressive multimodal understanding and generation},
  author={Zhao, Yue and Xue, Fuzhao and Reed, Scott and Fan, Linxi and Zhu, Yuke and Kautz, Jan and Yu, Zhiding and Kr{\"a}henb{\"u}hl, Philipp and Huang, De-An},
  journal={arXiv preprint arXiv:2502.05178},
  year={2025}
}

@article{chen2025blip3,
  title={Blip3-o: A family of fully open unified multimodal models-architecture, training and dataset},
  author={Chen, Jiuhai and Xu, Zhiyang and Pan, Xichen and Hu, Yushi and Qin, Can and Goldstein, Tom and Huang, Lifu and Zhou, Tianyi and Xie, Saining and Savarese, Silvio and others},
  journal={arXiv preprint arXiv:2505.09568},
  year={2025}
}

@article{tang2025unilip,
  title={Unilip: Adapting clip for unified multimodal understanding, generation and editing},
  author={Tang, Hao and Xie, Chenwei and Bao, Xiaoyi and Weng, Tingyu and Li, Pandeng and Zheng, Yun and Wang, Liwei},
  journal={arXiv preprint arXiv:2507.23278},
  year={2025}
}

@article{lin2025toklip,
  title={Toklip: Marry visual tokens to clip for multimodal comprehension and generation},
  author={Lin, Haokun and Wang, Teng and Ge, Yixiao and Ge, Yuying and Lu, Zhichao and Wei, Ying and Zhang, Qingfu and Sun, Zhenan and Shan, Ying},
  journal={arXiv preprint arXiv:2505.05422},
  year={2025}
}

@inproceedings{xie2025show,
  title={Show-o: One single transformer to unify multimodal understanding and generation},
  author={Xie, Jinheng and Mao, Weijia and Bai, Zechen and Zhang, David Junhao and Wang, Weihao and Lin, Kevin Qinghong and Gu, Yuchao and Chen, Zhijie and Yang, Zhenheng and Shou, Mike Zheng},
  booktitle={International Conference on Learning Representations},
  volume={2025},
  pages={28240--28264},
  year={2025}
}

@article{wan2025,
      title={Wan: Open and Advanced Large-Scale Video Generative Models},
      author={Team Wan and Ang Wang and Baole Ai and Bin Wen and Chaojie Mao and Chen-Wei Xie and Di Chen and Feiwu Yu and Haiming Zhao and Jianxiao Yang and Jianyuan Zeng and Jiayu Wang and Jingfeng Zhang and Jingren Zhou and Jinkai Wang and Jixuan Chen and Kai Zhu and Kang Zhao and Keyu Yan and Lianghua Huang and Mengyang Feng and Ningyi Zhang and Pandeng Li and Pingyu Wu and Ruihang Chu and Ruili Feng and Shiwei Zhang and Siyang Sun and Tao Fang and Tianxing Wang and Tianyi Gui and Tingyu Weng and Tong Shen and Wei Lin and Wei Wang and Wei Wang and Wenmeng Zhou and Wente Wang and Wenting Shen and Wenyuan Yu and Xianzhong Shi and Xiaoming Huang and Xin Xu and Yan Kou and Yangyu Lv and Yifei Li and Yijing Liu and Yiming Wang and Yingya Zhang and Yitong Huang and Yong Li and You Wu and Yu Liu and Yulin Pan and Yun Zheng and Yuntao Hong and Yupeng Shi and Yutong Feng and Zeyinzi Jiang and Zhen Han and Zhi-Fan Wu and Ziyu Liu},
      journal = {arXiv preprint arXiv:2503.20314},
      year={2025}
}

@inproceedings{chen2024internvl,
  title={Internvl: Scaling up vision foundation models and aligning for generic visual-linguistic tasks},
  author={Chen, Zhe and Wu, Jiannan and Wang, Wenhai and Su, Weijie and Chen, Guo and Xing, Sen and Zhong, Muyan and Zhang, Qinglong and Zhu, Xizhou and Lu, Lewei and others},
  booktitle={Proceedings of the IEEE/CVF conference on computer vision and pattern recognition},
  pages={24185--24198},
  year={2024}
}

@article{chen2025janus,
  title={Janus-pro: Unified multimodal understanding and generation with data and model scaling},
  author={Chen, Xiaokang and Wu, Zhiyu and Liu, Xingchao and Pan, Zizheng and Liu, Wen and Xie, Zhenda and Yu, Xingkai and Ruan, Chong},
  journal={arXiv preprint arXiv:2501.17811},
  year={2025}
}

@article{wu2025omnigen2,
  title={Omnigen2: Exploration to advanced multimodal generation},
  author={Wu, Chenyuan and Zheng, Pengfei and Yan, Ruiran and Xiao, Shitao and Luo, Xin and Wang, Yueze and Li, Wanli and Jiang, Xiyan and Liu, Yexin and Zhou, Junjie and others},
  journal={arXiv preprint arXiv:2506.18871},
  year={2025}
}

@article{xie2026show,
  title={Show-o2: Improved native unified multimodal models},
  author={Xie, Jinheng and Yang, Zhenheng and Shou, Mike Zheng},
  journal={Advances in Neural Information Processing Systems},
  volume={38},
  pages={47490--47518},
  year={2026}
}

@article{wang2025ovis,
  title={Ovis-u1 technical report},
  author={Wang, Guo-Hua and Zhao, Shanshan and Zhang, Xinjie and Cao, Liangfu and Zhan, Pengxin and Duan, Lunhao and Lu, Shiyin and Fu, Minghao and Chen, Xiaohao and Zhao, Jianshan and others},
  journal={arXiv preprint arXiv:2506.23044},
  year={2025}
}

@article{tian2026internvl,
  title={Internvl-u: Democratizing unified multimodal models for understanding, reasoning, generation and editing},
  author={Tian, Changyao and Yang, Danni and Chen, Guanzhou and Cui, Erfei and Wang, Zhaokai and Duan, Yuchen and Yin, Penghao and Chen, Sitao and Yang, Ganlin and Liu, Mingxin and others},
  journal={arXiv preprint arXiv:2603.09877},
  year={2026}
}

@inproceedings{liu2026tuna,
  title={Tuna: Taming unified visual representations for native unified multimodal models},
  author={Liu, Zhiheng and Ren, Weiming and Liu, Haozhe and Zhou, Zijian and Chen, Shoufa and Qiu, Haonan and Huang, Xiaoke and An, Zhaochong and Yang, Fanny and Patel, Aditya and others},
  booktitle={Proceedings of the IEEE/CVF Conference on Computer Vision and Pattern Recognition},
  pages={15740--15751},
  year={2026}
}

@article{lin2025uniworld,
  title={Uniworld-v1: High-resolution semantic encoders for unified visual understanding and generation},
  author={Lin, Bin and Li, Zongjian and Cheng, Xinhua and Niu, Yuwei and Ye, Yang and He, Xianyi and Yuan, Shenghai and Yu, Wangbo and Wang, Shaodong and Ge, Yunyang and others},
  journal={arXiv preprint arXiv:2506.03147},
  year={2025}
}

@article{li2025uniworld,
  title={Uniworld-v2: Reinforce image editing with diffusion negative-aware finetuning and mllm implicit feedback},
  author={Li, Zongjian and Liu, Zheyuan and Zhang, Qihui and Lin, Bin and Wu, Feize and Yuan, Shenghai and Yan, Zhiyuan and Ye, Yang and Yu, Wangbo and Niu, Yuwei and others},
  journal={arXiv preprint arXiv:2510.16888},
  year={2025}
}

@article{team2025longcat,
  title={Longcat-image technical report},
  author={Team, Meituan LongCat and Ma, Hanghang and Tan, Haoxian and Huang, Jiale and Wu, Junqiang and He, Jun-Yan and Gao, Lishuai and Xiao, Songlin and Wei, Xiaoming and Ma, Xiaoqi and others},
  journal={arXiv preprint arXiv:2512.07584},
  year={2025}
}

@article{cai2025z,
  title={Z-image: An efficient image generation foundation model with single-stream diffusion transformer},
  author={Cai, Huanqia and Cao, Sihan and Du, Ruoyi and Gao, Peng and Hoi, Steven and Hou, Zhaohui and Huang, Shijie and Jiang, Dengyang and Jin, Xin and Li, Liangchen and others},
  journal={arXiv preprint arXiv:2511.22699},
  year={2025}
}

@inproceedings{xiao2025omnigen,
  title={OmniGen: Unified Image Generation},
  author={Shitao Xiao and Yueze Wang and Junjie Zhou and Huaying Yuan and Xingrun Xing and Ruiran Yan and Chaofan Li and Shuting Wang and Tiejun Huang and Zheng Liu},
  booktitle={2025 IEEE/CVF Conference on Computer Vision and Pattern Recognition (CVPR)},
  year={2025}
}

@article{sun2024autoregressive,
  title={Autoregressive Model Beats Diffusion: Llama for Scalable Image Generation},
  author={Peize Sun and Yi Jiang and Shoufa Chen and Shilong Zhang and Bingyue Peng and Ping Luo and Zehuan Yuan},
  journal={arXiv preprint arXiv:2406.06525},
  year={2024}
}

@inproceedings{caron2021emerging,
  title={Emerging Properties in Self-Supervised Vision Transformers},
  author={Caron, Mathilde and Touvron, Hugo and Misra, Ishan and J\'egou, Herv\'e  and Mairal, Julien and Bojanowski, Piotr and Joulin, Armand},
  booktitle={Proceedings of the International Conference on Computer Vision (ICCV)},
  year={2021}
}
\bibliographystyle{iclr2026_conference}

\clearpage
\appendix
\section{Ablation on Balanced Flow Matching}

We ablate the reconstruction-component objective weight \(\lambda_r\) under the same \(768{+}128\) unified representation used by our final model. All variants share the same encoder, decoder, training schedule, and sampling setting, isolating the effect of the semantic--reconstruction balance in the flow-matching objective.

The sweep shows a clear middle optimum rather than a monotonic preference for
larger reconstruction weights. Increasing the reconstruction objective share
from $\lambda_r=0.25$ to $\lambda_r=0.75$ improves FID from $8.99$ to $7.10$,
indicating that under-weighting the reconstruction component leaves
decoder-critical visual details insufficiently modeled. However, further
increasing the reconstruction share does not continue to improve generation,
with $\lambda_r=0.92$ producing $7.39$. This supports assigning sufficient but
not overwhelming objective weight to the reconstruction component. The best
FID is obtained at $\lambda_r=0.75$, matching the weighting used in our final
balanced flow-matching objective.

\begin{table}[h]
\centering
\caption{Ablation of reconstruction-component weighting in balanced flow
matching. All variants use the same $768{+}128$ unified representation and are
evaluated under the same controlled generation setting.}
\label{tab:bfm-ablation}
\setlength{\tabcolsep}{7.0pt}
\renewcommand{\arraystretch}{1.12}
\begin{tabular}{cccc}
\toprule
Recon. share $\lambda_r$ & FID$\downarrow$ & IS$\uparrow$ & Precision$\uparrow$ \\
\midrule
$0.25$ & 8.99 & 142.4 & 0.681 \\
$0.60$ & 7.42 & 153.2 & 0.696 \\
$0.75$ & \textbf{7.10} & \textbf{154.4} & 0.704 \\
$0.82$ & 7.24 & 152.8 & \textbf{0.705} \\
$0.86$ & 7.13 & 152.5 & 0.704 \\
$0.92$ & 7.39 & 150.0 & 0.702 \\
\bottomrule
\end{tabular}
\end{table}

\section{Ablation on Reconstruction Compression}

We next study the channel compression in the Channel Factorized Merge. This
ablation compares our default $128$-dimensional reconstruction component with
a no-compression variant that preserves the full $768$ reconstruction
channels. Both variants use the same semantic component and the same
factorized concatenation design; only the dimensionality of
$\widetilde{T}_r$ is changed.

The no-compression variant confirms the expected reconstruction benefit of a
larger reconstruction stream: increasing $d_r$ from $128$ to $768$ improves
rFID from $0.163$ to $0.085$ and PSNR from $29.79$ to $33.61$. However, this
extra detail comes at a clear cost for generation. At both $20$ and $40$
epochs, the full-dimensional reconstruction component is substantially harder
for the DiT prior to model, with FID degrading from $9.95$ to $15.51$ at epoch
$20$ and from $6.92$ to $11.07$ at epoch $40$. These results support the role
of compression in the Channel Factorized Merge: it is not intended to
maximize reconstruction in isolation, but to retain sufficient
decoder-critical details while reducing the complexity of the reconstruction
distribution that the generative model must learn.

\begin{table}[h]
\centering
\caption{Ablation of reconstruction-component compression. Reducing the
reconstruction stream from $768$ to $128$ channels slightly weakens pure
reconstruction metrics, but substantially improves generative modeling under
the same early training budget.}
\label{tab:compression-ablation}
\setlength{\tabcolsep}{3.6pt}
\renewcommand{\arraystretch}{1.12}
\resizebox{\textwidth}{!}{%
\begin{tabular}{cccrrrrrrrr}
\toprule
Recon. dim $d_r$ & Unified dim & Epoch & PSNR$\uparrow$ & SSIM$\uparrow$ & rFID$\downarrow$ & FID$\downarrow$ & sFID$\downarrow$ & IS$\uparrow$ & Pre.$\uparrow$ & Rec.$\uparrow$ \\
\midrule
$128$ & $768{+}128$ & $20$ & 29.79 & 0.873 & 0.163 & 9.95 & 7.73 & 117.5 & 0.713 & 0.565 \\
$128$ & $768{+}128$ & $40$ & 29.79 & 0.873 & 0.163 & \textbf{6.92} & \textbf{6.93} & \textbf{144.4} & \textbf{0.730} & 0.587 \\
\midrule
$768$ & $768{+}768$ & $20$ & \textbf{33.61} & \textbf{0.940} & \textbf{0.085} & 15.51 & 8.45 & 90.0 & 0.655 & 0.574 \\
$768$ & $768{+}768$ & $40$ & \textbf{33.61} & \textbf{0.940} & \textbf{0.085} & 11.07 & 7.80 & 115.2 & 0.679 & \textbf{0.596} \\
\bottomrule
\end{tabular}%
}
\end{table}

\section{Entangled Representation and Diagnostic Protocol}
\label{app:entangled-diagnostic}

\label{app:entangled-diagnostic}

We provide additional details of the diagnostic experiment in
Sec.~\ref{sec:diagnostic-study}. Let $I$ denote an input image, and let
$$
Z_s=E_s(I), \qquad Z_r=E_r(I)
$$
denote the outputs of a pretrained semantic encoder and a
reconstruction-oriented encoder, respectively. We merge the two representations
using an MLP:
$$
Z_m=M([Z_s;Z_r]),
$$
where $[\cdot\,;\cdot]$ denotes channel-wise concatenation. Unlike the
factorized representation used in our method, $Z_m$ does not preserve an
explicit semantic--reconstruction decomposition.

The merger is trained with the same semantic and reconstruction alignment
objectives described in the main text. Specifically, $Z_m$ is aligned with
$Z_s$ to preserve semantic information, while a linear projection of $Z_m$ is
aligned with $Z_r$ to retain reconstruction-relevant information. The semantic
decoder $D_s$ and the high-fidelity reconstruction decoder $D_r$ are trained
separately before the diagnostic experiment, with their corresponding encoders
kept frozen. At the representation level, the resulting $Z_m$ simultaneously
supports semantic understanding and high-fidelity reconstruction on real
encoded latents. It achieves a zero-shot accuracy of $78.53$, compared with
$79.10$ for the SigLIP baseline, together with a PSNR of $33.83$ and an rFID
of $0.069$.

We next train a DiT prior directly on $Z_m$ for $20$ epochs using the same
image distribution and evaluation protocol as in the main experiments. The
resulting generated latent exhibits a pronounced decoder-dependent gap:
decoding it with the high-fidelity reconstruction decoder $D_r$ yields an FID
of $120.9$, whereas decoding it with the semantic decoder $D_s$ yields a much
lower FID of $8.07$. This gap indicates that the DiT prior primarily learns
the semantic-dominant variation in the entangled representation, while failing
to model the reconstruction-relevant variation required by $D_r$. Thus, the
fact that $Z_m$ supports both semantic understanding and high-fidelity
reconstruction on real encoded latents does not ensure that both types of
information are equally accessible to a generative prior. The reconstruction
information is present, but its entangled composition with semantic variation
makes it difficult for the generative model to identify and control.

To quantify the relative contribution of the semantic pathway, we measure the
fraction of the merger output variation explained by the semantic input:
$$
\rho_s
=
\frac{\mathrm{Var}(M(Z_s,0))}
{\mathrm{Var}(M(Z_s,Z_r))}.
$$
The variance is computed over the evaluation image set using the same merger
parameters. We obtain $\rho_s\approx95\%$, indicating that most of the merger
output variation is explained by the semantic pathway. This suggests that
reconstruction-relevant directions, although sufficient for reconstructing real
encoded latents, occupy a relatively small and difficult-to-control part of
the entangled latent space. This observation motivates the explicit
factorization used in our final representation, where $T_s$ and
$\widetilde{T}_r$ are concatenated along the channel dimension and remain
directly accessible during flow-matching training.

\end{document}